\pdfoutput=1

\documentclass[11pt]{article}

\usepackage{acl}

\usepackage{times}
\usepackage{latexsym}

\usepackage[T1]{fontenc}

\usepackage[utf8]{inputenc}

\usepackage{microtype}

\usepackage{inconsolata}

\usepackage{algorithm}
\usepackage{algorithmic}
\usepackage{graphicx}
\usepackage{subfigure}
\usepackage{xfrac}
\usepackage{colortbl}
\usepackage{booktabs}  
\usepackage{amsfonts}  
\usepackage{nicefrac}
\usepackage{latexsym}
\usepackage{multirow}
\usepackage{amsmath}
\usepackage{tabularx}
\usepackage{makecell}
\usepackage{xcolor}
\usepackage{wrapfig}
\usepackage{enumerate}
\usepackage{engord}

\usepackage{tcolorbox}
\tcbuselibrary{theorems}
\newtcbtheorem[number within=section]{mytheo}{My Theorem}%
{colback=green!5,colframe=green!35!black,fonttitle=\bfseries}{th}

\usepackage{adjustbox}
\usepackage{tikz}
\usepackage{pgfplots}
\pgfplotsset{compat=1.18}
\usepackage[edges]{forest}
\definecolor{framework-blue}{RGB}{47, 85, 151}

\definecolor{content-yellow}{RGB}{255, 230, 153}
\definecolor{framework-yellow}{RGB}{255, 255, 255}
\definecolor{content-orange}{RGB}{251, 229, 215}
\definecolor{framework-orange}{RGB}{248, 203, 175}
\definecolor{content-gray}{RGB}{237, 237, 237}
\definecolor{framework-gray}{RGB}{166, 166, 166}

\definecolor{paired-light-blue}{RGB}{198, 219, 239}
\definecolor{paired-dark-blue}{RGB}{49, 130, 188}
\definecolor{paired-light-orange}{RGB}{251, 208, 162}
\definecolor{paired-dark-orange}{RGB}{230, 85, 12}
\definecolor{paired-light-green}{RGB}{199, 233, 193}
\definecolor{paired-dark-green}{RGB}{49, 163, 83}
\definecolor{paired-light-purple}{RGB}{218, 218, 235}
\definecolor{paired-dark-purple}{RGB}{117, 107, 176}
\definecolor{paired-light-gray}{RGB}{217, 217, 217}
\definecolor{paired-dark-gray}{RGB}{99, 99, 99}
\definecolor{paired-light-pink}{RGB}{222, 158, 214}
\definecolor{paired-dark-pink}{RGB}{123, 65, 115}
\definecolor{paired-light-red}{RGB}{231, 150, 156}
\definecolor{paired-dark-red}{RGB}{131, 60, 56}
\definecolor{paired-light-yellow}{RGB}{231, 204, 149}
\definecolor{paired-dark-yellow}{RGB}{141, 109, 49}
\tikzset{%
    parent/.style = {align=center,text width=2.5cm,rounded corners=3pt, line width=0.3mm, fill=gray!10,draw=gray!80},
    child/.style = {align=center,text width=2.3cm,rounded corners=3pt, fill=blue!10,draw=blue!80,line width=0.3mm},
    grandchild/.style = {align=center,text width=2cm,rounded corners=3pt},
    greatgrandchild/.style = {align=center,text width=1.5cm,rounded corners=3pt},
    greatgrandchild2/.style = {align=center,text width=1.5cm,rounded corners=3pt},    
    referenceblock/.style =  {align=center,text width=1.5cm,rounded corners=2pt},
    domain_box/.style= {align=center,text width=2.2cm,rounded corners=3pt, fill=white,draw=framework-blue,line width=0.3mm},
    datasets/.style= {align=center, text width=4.5cm,rounded corners=3pt, fill=paired-light-blue!45,draw=framework-blue,line width=0.3mm},
    few_datasets/.style= {align=center, text width=2.2cm,rounded corners=3pt, fill=paired-light-blue!45,draw=framework-blue,line width=0.3mm},
    models/.style= {align=center, text width=4.5cm,rounded corners=3pt, fill=paired-light-orange!45,draw=framework-blue,line width=0.3mm},
    few_models/.style= {align=center, text width=2.2cm,rounded corners=3pt, fill=paired-light-orange!45,draw=framework-blue,line width=0.3mm},
}

\newcommand{\ours}{\textsc{HD-Eval}}
\newcommand{\ourswb}{\textsc{HD-Eval} }

\title{\ours: Aligning Large Language Model Evaluators Through Hierarchical Criteria Decomposition}

\author{Yuxuan Liu\textsuperscript{$\dagger$}\thanks{~Work done during internship at Microsoft.}, Tianchi Yang\textsuperscript{$\ddagger$}, Shaohan Huang\textsuperscript{$\ddagger$}, Zihan Zhang\textsuperscript{$\ddagger$}, Haizhen Huang\textsuperscript{$\ddagger$}, \\
\textbf{Furu Wei\textsuperscript{$\ddagger$}, Weiwei Deng\textsuperscript{$\ddagger$}, Feng Sun\textsuperscript{$\ddagger$}, Qi Zhang\textsuperscript{$\ddagger$}}
\\
\textsuperscript{$\dagger$} Peking University \textsuperscript{$\ddagger$} Microsoft Corporation
\\
\small{\texttt{yx.liu@stu.pku.edu.cn}}
}

\begin{document}
\maketitle
\begin{abstract}
Large language models (LLMs) have emerged as a promising alternative to expensive human evaluations. However, the alignment and coverage of LLM-based evaluations are often limited by the scope and potential bias of the evaluation prompts and criteria. To address this challenge, we propose \ours, a novel framework that iteratively aligns LLM-based evaluators with human preference via \textbf{\underline{H}}ierarchical Criteria \textbf{\underline{D}}ecomposition. \ourswb inherits the essence from the evaluation mindset of human experts and enhances the alignment of LLM-based evaluators by decomposing a given evaluation task into finer-grained criteria, aggregating them according to estimated human preferences, pruning insignificant criteria with attribution, and further decomposing significant criteria. By integrating these steps within an iterative alignment training process, we obtain a hierarchical decomposition of criteria that comprehensively captures aspects of natural language at multiple levels of granularity. Implemented as a white box, the human preference-guided aggregator is efficient to train and more explainable than relying solely on prompting, and its independence from model parameters makes it applicable to closed-source LLMs. Extensive experiments on three evaluation domains demonstrate the superiority of \ourswb in further aligning state-of-the-art evaluators and providing deeper insights into the explanation of evaluation results and the task itself.
\end{abstract}

\section{Introduction}
\label{ch:intro}
With the rapid development of LLMs and rising significance on NLG evaluations, an emerging line of works explores utilizing LLM as reference-free text quality evaluators \cite{kocmi2023gemba, wang2023chatgptgood, fu2023gptscore, liu2023gpteval}. To leverage the instruction following capability of LLMs, existing works utilize a \textit{single} piece of criteria (as a prompt) to evaluate a given sample. Given the superior instruction-following capability and immense knowledge obtained through pre-training, LLM-based evaluators substantially outperform previous automatic evaluation metrics \cite{yuan2021bartscore, zhong2022unieval}, and opens a promising alternative for human evaluation.

However, despite their achievements, an emerging line of research questions the alignment and trustworthiness of LLM judgments. 
As recent studies point out, these approaches are limited by the bias of prompt design \cite{wang2023chatgptgood}, resulting in potential biases in its judgments \cite{wang2023notfair}, demanding per-task calibration on evaluation prompts to mitigate \cite{liu2023calibrating}. 

\begin{figure*}[t]
  \centering
  \includegraphics[width=\textwidth]{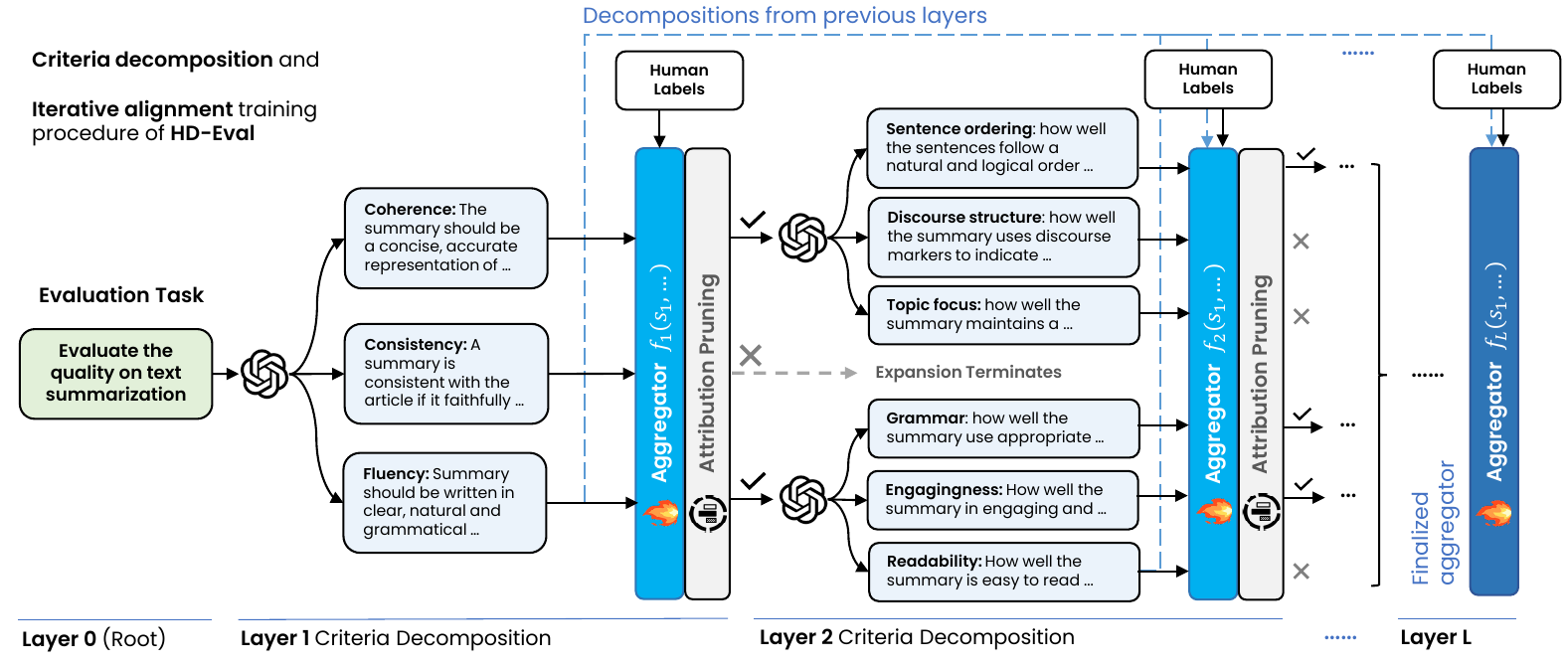}
  \caption{Overall framework of \ours. Starting from the evaluation task, \ourswb iteratively \textit{decomposes} it to different aspects, \textit{trains} an aggregator, then \textit{select} significant criteria with attribution pruning for further expansion at the next layer. The aggregator and decomposition are finalized after reaching the maximum layer count.}
  \label{fig:overview}
\end{figure*}

One core limitation of using a single criterion to evaluate text quality is that it may not capture the complexity and diversity of human evaluations and judgments. Human thinking is not linear or monolithic, but rather comprehensive and naturally follows a hierarchical order \cite{Tversky1974JudgmentUU}. When we read a book, we may evaluate it from different perspectives, such as plot, characters, style, and theme, each of which can further be naturally divided into more specific criteria. 

Hierarchical thinking \cite{haupt2018hierarchical} allows humans to resolve complex problems by first breaking them down into more tangible sub-problems, and then integrating the solutions at different levels of abstraction \cite{buzan2006mind}. Correspondingly, mainstream human evaluation protocols also leverage hierarchical critiques \cite{freitag2021experts}.

Our core motivation is to empower the alignment of LLM-based evaluators by rooting the evaluation mindset of human experts into design, while also harnessing state-of-the-art generic capabilities of LLMs. Drawing inspirations from the above, we propose \ours, a novel framework to align LLM-based evaluator towards human preference through \textbf{\underline{H}}ierarchical Criteria \textbf{\underline{D}}ecomposition. 

Specifically, the design of critical components of \ourswb inherits the essence of the human evaluation mindset: task decomposition, analysis of all sub-tasks, and a final comprehensive evaluation. Correspondingly, we propose 3 crucial stages: (1) \textit{Hierarchical Criteria Decomposition}, where we decompose an evaluation task into a hierarchy of evaluation criteria, each focusing on different evaluation aspects with various granularity; (2) \textit{Human Preference-Guided Aggregation}, where we aggregate evaluation results at each hierarchy to obtain a final judgment, with respect to the estimated preference of human experts on different hierarchies; (3) \textit{Attribution Pruning}, to dynamically attribute human expert's preference on existing criteria to efficiently prune the space of decomposition, focus on significant aspects, thus improving its fidelity. 

To align an LLM-based evaluator toward human preference, we propose \textit{Iterative Alignment Training Framework} to seamlessly integrate the 3 stages above in a layer-wise iterative fashion. When the training process of \ourswb completes, we obtain a pair of finalized criteria decomposition and human preference-guided aggregator, which could be applied to evaluation samples upon application. 

We highlight the following key contributions of \ourswb as follows:
\begin{enumerate}[1)]
  \itemsep0em
  \item We propose \ours, a novel framework that aligns LLM-based evaluators towards human preference via comprehensively decomposing criteria into multiple levels of hierarchy. 
  \item Implemented as white-box, judgments made by aggregators of \ourswb are more controllable and explainable than solely prompting LLMs. 
  \item The design of \ourswb ensures its applicability to both open-source and API-hosted LLMs.
 \item Comprehensive experiments on three evaluation domains demonstrate the superior capability of \ourswb in aligning LLM-based evaluators. 
\end{enumerate}

\section{Methodology}
\label{ch:method}

\begin{figure*}[ht]
  \centering
  \includegraphics[width=\textwidth]{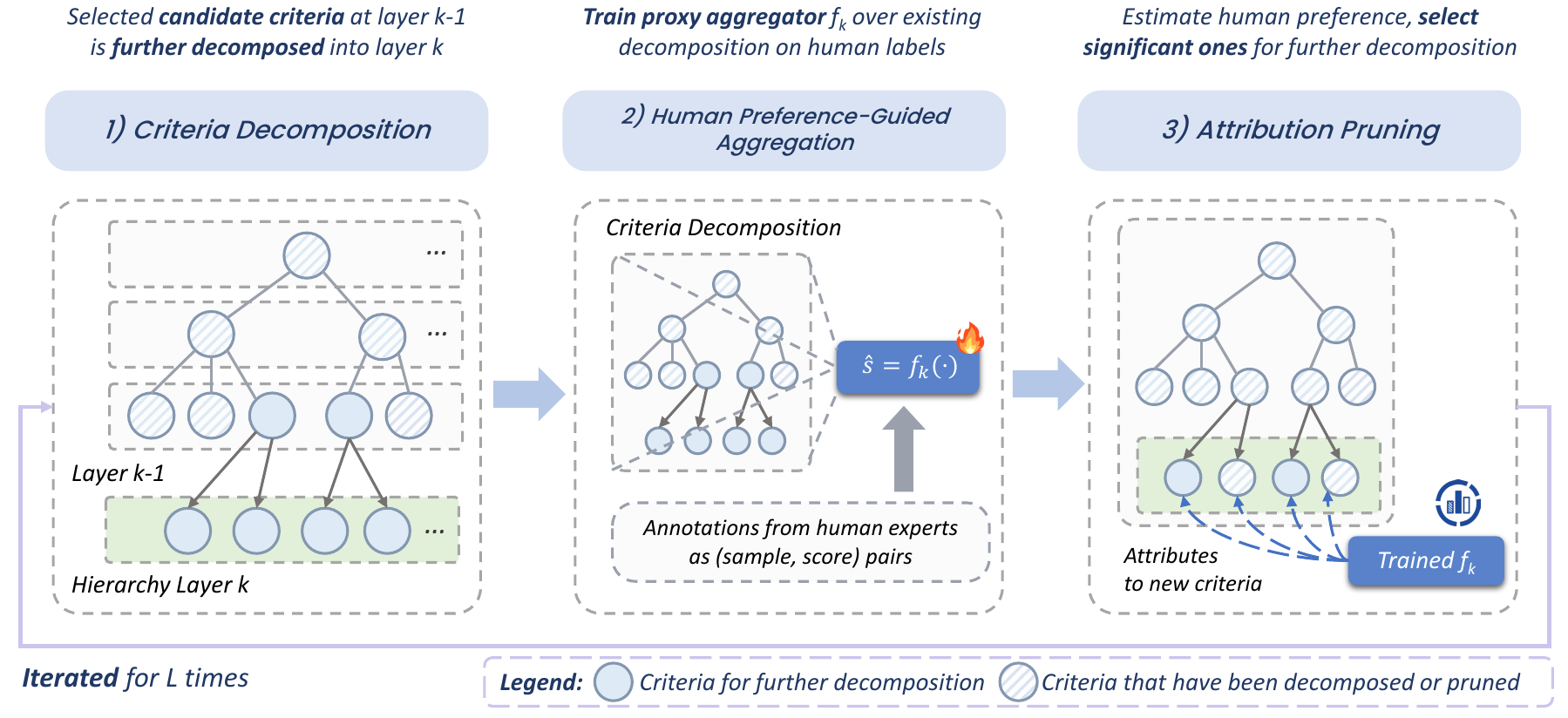}
  \caption{Illustration on hierarchical criteria decomposition and iterative alignment training of \ours. A formal description of the iterative alignment training procedure of \ourswb is elaborated in Algorithm \ref{alg:mainalgo}.}
  \label{fig:iter}
\end{figure*}

\subsection{Hierarchical Criteria Decomposition}
\label{sec:decomp}
To leverage the hierarchical thinking of human evaluation mindset and mitigate potential bias, we propose Hierarchical Criteria Decomposition to obtain a \textit{hierarchy} of evaluation criteria. This analogy of human evaluation mindset naturally reciprocates an \textit{alignment} between LLMs and expert evaluations.

\paragraph{Criteria Decomposition with LLMs} As illustrated in Figure \ref{fig:overview}, \ourswb iteratively decomposes an evaluation task into a hierarchy of criteria.
To obtain such decomposition, we prompt LLMs to obtain a decomposition of a single criteria, by providing backgrounds of the evaluation task $\mathcal{T}$ and the parent evaluation criteria $\mathcal{C}^{l-1}_j$:
\begin{equation}
\label{score}
    \{\mathcal{C}^{l}_1, ..., \mathcal{C}^{l}_m\} = LLM(\mathcal{T}, \mathcal{C}^{l-1}_j),
\end{equation}
where the $j$-th evaluation criteria at hierarchy level $l-1$ is further decomposed into a series of sub-criteria $\{\mathcal{C}^{l}_1, ..., \mathcal{C}^{l}_m\}$ by the LLM. By iteratively performing this decomposition starting from the overall task as \textit{root} node, we naturally obtain a tree-structured hierarchy of evaluation criteria, focusing on different evaluation levels and aspects.

\paragraph{Hierarchy-Aware Prompting} To leverage the hierarchical decomposition of criteria, we propose Hierarchy-Aware Prompting to preserve the hierarchical relations when evaluating a decomposed criteria (node). Specifically, when evaluating a single aspect \textit{(child)}, we also provide information from its \textit{parent} node. 
This prompt design reserves the local hierarchical information (i.e., \textit{links}), while refrains excessive and irrelevant information, providing LLMs a better grasp of the criteria\footnote{Full prompts are provided in Appendix \ref{app:prompts}.}.

\subsection{Human Preference-Guided Aggregation}
\label{sec:aggregate}
After obtaining decomposed sub-criteria from parent criteria with \ours, we propose Human Preference-Guided Aggregation to adequately address the importance of each decomposed criteria to obtain a final verdict. 
A concurrent work on decomposition \cite{saha2023branch} prompt the LLM itself for such verdict. However, it potentially suffers from the inherent bias of LLMs \cite{wang2023notfair}, and also fail to address \textit{human preference}.

To overcome these limitations, we adapt white-box aggregator to \textit{estimate} how human experts value each decomposed criteria. The aggregator $f_\theta$ serves as a human preference estimator to aggregate scores on different sub-criteria for comprehensive evaluation. The aggregator is trained as a regressor $f_\theta : \mathbb{R}^{|\mathcal{C}|} \to \mathbb{R}^{p}$, to map evaluation scores from decomposed criteria to human expert scores\footnote{Human score label ($s_k \in \mathbb{R}^p$) is a numeric vector containing evaluation scores for a total of $p$ evaluation aspects.} for a training sample, using MSE objective:
\begin{equation}
\label{eq:agg}
    \hat{s}_k=f_\theta(a^{1,1}_k, ..., a^{1,n}_k, ..., a^{L,1}_k, ..., a^{L,m}_k),
\end{equation}
where $a^{i,j}_k$ denotes the evaluation score (ranged in 0-5) for the $j$-th criteria of the $i$-th layer to sample $k$ ($\hat{s}_k\in\mathbb{R}^p$). Equation \ref{eq:agg} essentially learns to assign attention to scores from different decomposed criterion, which is in equal to implicitly estimating how human experts value each decomposed criterion.

\renewcommand\arraystretch{1.1}
\begin{table*}[t]
\center \scriptsize
\tabcolsep0.1 in
\resizebox{0.99\textwidth}{!}{

\begin{tabular}{lcccccccccc}
\toprule
\multicolumn{1}{l}{\multirow{2}[1]{*}{\textbf{Metrics}}} & \multicolumn{2}{c}{\textbf{Coherence}}
 & \multicolumn{2}{c}{\textbf{Consistency}} & \multicolumn{2}{c}{\textbf{Fluency}} & \multicolumn{2}{c}{\textbf{Relevance}} & \multicolumn{2}{c}{\textbf{Average}} \\
 & \multicolumn{1}{c}{$r$} & \multicolumn{1}{c}{$\rho$} & \multicolumn{1}{c}{$r$} & \multicolumn{1}{c}{$\rho$} & \multicolumn{1}{c}{$r$} & \multicolumn{1}{c}{$\rho$} & \multicolumn{1}{c}{$r$} & \multicolumn{1}{c}{$\rho$} & \multicolumn{1}{c}{$r$} & \multicolumn{1}{c}{$\rho$}  \\
 
\cmidrule(lr){1-1} \cmidrule(lr){2-3} \cmidrule(lr){4-5} \cmidrule(lr){6-7} \cmidrule(lr){8-9} \cmidrule(lr){10-11}

\textsc{ROUGE-1} & 0.178 & 0.168 & 0.037 & 0.028 & 0.045 & 0.009 & 0.288 & 0.291 & 0.137 & 0.124 \\
\textsc{ROUGE-2} & 0.143 & 0.152 & 0.025 & 0.011 & 0.029 & -0.006 & 0.209 & 0.240 & 0.101 & 0.099 \\
\textsc{ROUGE-L} & 0.141 & 0.134 & 0.026 & 0.015 & 0.052 & 0.022 & 0.262 & 0.264 & 0.120 & 0.109 \\
\textsc{BertScore} & 0.302 & 0.285 & 0.093 & 0.071 & 0.174 & 0.119 & 0.389 & 0.372 & 0.239 & 0.212 \\

\textsc{PRISM} & 0.188 & 0.184 & 0.067 & 0.039 & 0.074 & 0.053 & 0.290 & 0.290 & 0.154 & 0.141 \\

\textsc{CTC} & 0.220 & 0.181 & 0.531 & 0.407 & 0.494 & 0.305 & 0.259 & 0.127 & 0.376 & 0.255 \\

\textsc{BartScore} & 0.423 & 0.403 & 0.350 & 0.317 & 0.303 & 0.250 & 0.415 & 0.386 & 0.373 & 0.339 \\

\textsc{UniEval} & 0.545 & 0.588 & 0.602 & 0.439 & 0.601 & 0.460 & 0.464 & 0.478 & 0.553 & 0.491 \\


\textsc{GPT-4 Eval} & 0.547 & 0.542 & 0.507 & 0.458 & 0.479 & 0.460 & 0.609 & 0.592 & 0.538 & 0.513 \\ \midrule
\multicolumn{11}{c}{\cellcolor[rgb]{ .886,  .937,  .855} \textit{Iterative alignment training on $\textbf{25}\%$ of all human expert preference data}} \\
\addlinespace[0.75ex]
\ours\textsc{-NN} & 0.655 & 0.644 & 0.573 & 0.457 & 0.562 & 0.437 & 0.601 & 0.577 & 0.598 & 0.529 \\
\midrule

\multicolumn{11}{c}{\cellcolor[rgb]{ .886,  .937,  .855} \textit{Iterative alignment training on $\textbf{50}\%$ of all human expert preference data}} \\
\addlinespace[0.75ex]
\ours\textsc{-NN} & 0.668 & 0.657 & 0.604 & 0.451 & 0.580 & 0.435 & 0.619 & 0.599 & 0.617 & 0.535 \\
\bottomrule
\end{tabular}
}

\caption{Segment-level Pearson ($r$) and Spearman ($\rho$) human correlations of aspects on SummEval.}
\label{tab:sme}

\end{table*}

\subsection{Attribution Pruning}
\label{sec:pruning} The core motivation for attribution pruning is to ensure most searching efforts (i.e., \textit{deeper} decomposition) are focused on the most significant evaluation aspects. While it is feasible to obtain a \textit{full} tree-like hierarchical decomposition, it brings higher costs and might potentially introduce noisy or redundant criteria. However, it is non-trivial to assign importance to each generated criteria, as it demands domain expertise from human experts.

To remedy the demand on domain expertise, we propose Attribution Pruning to \textit{objectively} select the most significant criteria and further support it with augmented evidence, through continuing decomposing it into finer-grained criteria. 
As illustrated in Figure \ref{fig:overview}, after generating a new sub-criteria sets $C_i$ at the $i$-th iteration, we train a proxy aggregator $f_i(\cdot)$ to approximate human expert's preference on newly generated criteria.
Through training $f_i(\cdot)$, the human preference of each sub-criteria to the final verdict is implicitly assigned, which could be quantitatively measured with a saliency function $g(\cdot)$, with which we objectively attribute then select a \textit{significant subset} of criteria within $C_i$ to further decompose at the $i+1$-th iteration (we denote this subset as $C_D^{i+1}$):
\begin{equation}
    \mathcal{C}_D^{i+1} = \mathrm{argtop}k_{\mathcal{C}_D \in \mathcal{C}_i} \left[g\left(f_i(\mathcal{C})\right)\right],
\end{equation}
where $\mathcal{C}=\cup_i \mathcal{C}_i$ denote existing criteria set, and $k$ controls the maximum count of new criteria\footnote{Since criteria on upper levels are already being decomposed, we only select $\mathcal{C}_D^{i+1}$ within $\mathcal{C}_{i}$.}.
Since $f_i(\cdot)$ is a white-box, $g(\cdot)$ could be implemented as attribution methods\footnote{We adapt \textit{permutation importance} in this paper, since criteria whose score has larger permutation importance are more crucial to making the final comprehensive judgement.} (e.g., permutation importance \cite{altmann2010permutation}, Shapley additive explanations \citep{lundberg2017unified}), which provides superior controllability and explainability, compared to prompting or tuning of LLMs.

\subsection{Iterative Alignment Training Framework}
\label{sec:iter_train}
Combining the above, we propose an Iterative Alignment Training Framework for \ours, as summarized in Figure \ref{fig:iter}. We seamlessly integrate critical components, i.e. criteria decomposition, human preference-guided aggregation, and attribution pruning in a per-layer iterative fashion. 

Specifically, In $j$-th training iteration, we first perform criteria decomposition to each of criteria in candidates $\mathcal{C}_D^{j}$ selected from the last step with pruning, obtaining a set of new criteria $\mathcal{C}_{j}$ for $j$-th layer. We then train a new proxy aggregator $f_j(\cdot)$ to estimate human preference, and finally perform attribution pruning based on $f_j(\cdot)$ to select significant criteria $\mathcal{C}_D^{j+1}$ for decomposition at the next iteration. When this iterative alignment training completes, we obtain a pair of \textit{finalized} aggregator and criteria decomposition, which could be applied to new candidate evaluation samples upon application. The exact learning process of \ourswb is formally summarized in Algorithm \ref{alg:mainalgo}.

\section{Experiments}
\label{ch:exp}
\renewcommand\arraystretch{1.1}
\begin{table*}[t]
\center \scriptsize
\tabcolsep0.1 in
\resizebox{0.99\textwidth}{!}{

\begin{tabular}{lcccccccccc}
\toprule
\multicolumn{1}{l}{\multirow{2}[1]{*}{\textbf{Metrics}}} & \multicolumn{2}{c}{\textbf{Naturalness}}
 & \multicolumn{2}{c}{\textbf{Coherence}} & \multicolumn{2}{c}{\textbf{Engagingness}} & \multicolumn{2}{c}{\textbf{Groundedness}} & \multicolumn{2}{c}{\textbf{Average}} \\
 & \multicolumn{1}{c}{$r$} & \multicolumn{1}{c}{$\rho$} & \multicolumn{1}{c}{$r$} & \multicolumn{1}{c}{$\rho$} & \multicolumn{1}{c}{$r$} & \multicolumn{1}{c}{$\rho$} & \multicolumn{1}{c}{$r$} & \multicolumn{1}{c}{$\rho$} & \multicolumn{1}{c}{$r$} & \multicolumn{1}{c}{$\rho$}  \\
 
\cmidrule(lr){1-1} \cmidrule(lr){2-3} \cmidrule(lr){4-5} \cmidrule(lr){6-7} \cmidrule(lr){8-9} \cmidrule(lr){10-11}

\textsc{ROUGE-1} & 0.158 & 0.143 & 0.205 & 0.206 & 0.305 & 0.319 & 0.264 & 0.264 & 0.233 & 0.233 \\
\textsc{ROUGE-2} & 0.175 & 0.168 & 0.186 & 0.247 & 0.281 & 0.337 & 0.260 & 0.311 & 0.225 & 0.266 \\
\textsc{ROUGE-L} & 0.172 & 0.145 & 0.198 & 0.205 & 0.299 & 0.306 & 0.286 & 0.293 & 0.239 & 0.237 \\
\textsc{BertScore} & 0.226 & 0.209 & 0.214 & 0.233 & 0.317 & 0.335 & 0.291 & 0.317 & 0.262 & 0.273\\

\textsc{PRISM} & 0.040 & -0.010 & 0.098 & 0.081 & 0.241 & 0.220 & 0.178 & 0.159 & 0.139 & 0.113 \\

\textsc{CTC} & 0.232 & 0.195 & 0.343 & 0.296 & 0.540 & 0.542 & 0.422 & 0.398 & 0.384 & 0.358 \\

\textsc{BartScore} & -0.072 & -0.053 & -0.107 & -0.079 & -0.105 & -0.084 & -0.217 & -0.197 & -0.125 & -0.103 \\

\textsc{UniEval} & 0.342 & 0.450 & 0.571 & 0.616 & 0.573 & 0.615 & 0.523 & 0.590 & 0.502 & 0.568 \\


\textsc{GPT-4 Eval} & 0.584 & 0.607 & 0.562 & 0.590 & 0.594 & 0.605 & 0.530 & 0.556 & 0.567 & 0.590 \\ \midrule
\multicolumn{11}{c}{\cellcolor[rgb]{ .886,  .937,  .855} \textit{Iterative alignment training on $\textbf{25}\%$ of all human expert preference data}} \\
\addlinespace[0.75ex]
\ours\textsc{-NN} & 0.647 & 0.672 & 0.588 & 0.613 & 0.682 & 0.702 & 0.471 & 0.498 & 0.597 & 0.621 \\
\midrule

\multicolumn{11}{c}{\cellcolor[rgb]{ .886,  .937,  .855} \textit{Iterative alignment training on $\textbf{50}\%$ of all human expert preference data}} \\
\addlinespace[0.75ex]
\ours\textsc{-NN} & 0.648 & 0.674 & 0.584 & 0.607 & 0.682 & 0.701 & 0.549 & 0.568 & 0.616 & 0.638 \\

\bottomrule
\end{tabular}
}

\caption{Turn-level Pearson ($r$) and Spearman ($\rho$) human correlations of aspects on Topical-Chat.}
\label{tab:tpchat}

\end{table*}

\subsection{Experimental Setup}
\label{ch:exp_setup}
\paragraph{Datasets and Evaluations} We evaluate the performance of \ourswb on three NLG evaluation scenario: Summarization (SummEval \cite{fabbri2021summeval}), Conversation (Topical-Chat \cite{DBLP:conf/interspeech/GopalakrishnanH19}) and Data-to-Text (SFRES and SFHOT \cite{wen2015sfressfhot}). For assessing human alignment, we report dataset (segment) level meta-evaluation results on both Pearson's $r$ and Spearman's $\rho$ coefficient with human annotations. For each dataset, a $50\%$ proportion is held out for testing, while the rest is applied for training\footnote{We explore utilizing different percentages of training data in our experiments. Detailed count of training data will be reported under different experimental settings.}. 

\paragraph{Baselines} We compare our \ourswb against a series of automatic evaluation baselines, including ROUGE \citep{lin2004rouge}, BERTScore \citep{Zhang*2020BERTScore:},  PRISM \citep{thompson2020prism}, BartScore \citep{yuan2021bartscore}, and UniEval \citep{zhong2022unieval}. For LLM-based evaluation, we select \textsc{GPT-4} Evaluation \cite{liu2023gpteval}, representing state-of-the-art LLM-based evaluators.

\paragraph{Models and Configurations} We adopt OpenAI's GPT-4 model \cite{OpenAI2023GPT4TR} (\texttt{GPT-4-32K}) and LLama-2 families \citep{touvron2023llama}\footnote{Comprehensive studies on Llama-based \ourswb are presented in Appendix \ref{apd:llama} due to space limitations.} as LLM in this study. For the aggregator, we experiment with multiple white-box implementations, including Linear Regression (LR), Decision Tree (DT), Random Forest (RF), and shallow MLPs (NN). For criteria decomposition, we apply a maximum layer of $3$, and a child count of $4$ for parent nodes. Detailed implementations are listed in Appendix \ref{app:impl}.

\subsection{Experimental Results}
\label{ch:results}

\paragraph{Human Alignment} Meta evaluation results for \ourswb on evaluating summarization is listed in Table \ref{tab:sme}. We train our \ourswb under two data settings, representing \ourswb data and/or resource-constraint scenarios. As illustrated in Table \ref{tab:sme}, \ourswb substantially improved the human relevance of evaluation over GPT-4, resulting in a $15\%$ improvement on Pearson's correlation overall, and over $20\%$ in coherence and fluency. When training with only half of human expert annotations, the performance of \ourswb remains on-par or marginally off, demonstrating the effectiveness of the iterative alignment training process.

Similarly, in evaluating natural language conversations (Table \ref{tab:tpchat}), \ourswb improves the alignment of GPT-4 by uplifting both the Pearson and Spearman correlation over $8\%$, and maintained on-par performance on 3 of 4 evaluation aspects when training with only half of human preference data.

We finally test \ourswb on a more challenging evaluation task, i.e. evaluating the naturalness of data-to-text generations. As illustrated in Table \ref{tab:qags}, \ourswb obtained more than $15\%$ improvement in human correlations on both correlation coefficients and only lost around $3\%$ performance with only half of the training data available. These results highlight the effectiveness and efficiency of \ourswb in aligning LLM-based evaluators.

\begin{table}[t]
\center 

\tabcolsep0.07 in

\resizebox{0.49\textwidth}{!}{
\begin{tabular}{lcccccc}
\toprule
\multicolumn{1}{l}{\multirow{2}[1]{*}{\textbf{Metrics}}} & \multicolumn{2}{c}{\textbf{SFRES}}
 & \multicolumn{2}{c}{\textbf{SFHOT}} & \multicolumn{2}{c}{\textbf{Average}}  \\
 
 & $r$ & $\rho$ &  $r$ & $\rho$ & $r$ & $\rho$ \\
 
\cmidrule(lr){1-1} \cmidrule(lr){2-3} \cmidrule(lr){4-5} \cmidrule(lr){6-7}

\textsc{ROUGE-1} & 0.074 & 0.092 & 0.035 & 0.031 &   0.055 &   0.062  \\

\textsc{ROUGE-2}  & 0.094 & 0.073  & 0.060 & 0.042 &   0.077 &   0.051  \\

\textsc{ROUGE-L}  & 0.059 & 0.067 & 0.048 & 0.038 &   0.063 &   0.043  \\

\textsc{BertScore} & 0.164 & 0.145  & 0.103 & 0.087 &   0.134 &   0.116  \\

\textsc{PRISM}  & 0.146 & 0.126 & 0.164 & 0.131 &   0.155 &   0.129  \\

\textsc{BartScore}  & 0.280 & 0.255 & 0.133 & 0.095 &   0.207 &   0.175  \\

\textsc{CTC}  & 0.100 & 0.086 & 0.181 & 0.160 &   0.141 &   0.123  \\

\textsc{UniEval} & 0.381 & 0.354 & 0.350 & 0.305 &  0.366 &   0.330  \\

\textsc{GPT-4 Eval} & 0.414 & 0.347 & 0.436 & 0.364 & 0.425 & 0.356 \\ \midrule

\multicolumn{7}{c}{\cellcolor[rgb]{ .886,  .937,  .855} \textit{Iterative alignment training on $\textbf{25}\%$ of data}} \\
\addlinespace[0.5ex]
\ours\textsc{-NN} & 0.453 & 0.363 & 0.494 & 0.420 & 0.474 & 0.392 \\
\midrule

\multicolumn{7}{c}{\cellcolor[rgb]{ .886,  .937,  .855} \textit{Iterative alignment training on $\textbf{50}\%$ of data}} \\
\addlinespace[0.5ex]
\ours\textsc{-NN} & 0.470 & 0.389 & 0.510 & 0.432 & 0.490 & 0.411 \\
\bottomrule
\end{tabular}
}
\caption{Segment-level Pearson ($r$) and Spearman ($\rho$) correlations on 
Data-to-Text generation tasks.}
\label{tab:qags}
\end{table}

\begin{figure*}[t] 
\scriptsize 
\begin{adjustbox}{width=\textwidth} 
    \begin{forest} 
            for tree={
                forked edges,
                grow'=0,
                draw,
                rounded corners,
                node options={align=center},
                text width=2.7cm,
                s sep=0.25pt,
                calign=edge midpoint, 
            }, 
            [\textit{Target NLG Evaluation Task}: Evaluate the quality of natural language conversation generations, fill=gray!45, parent 
                [Naturalness (\textcolor{violet}{\textit{nat}}), domain_box [Grammar and syntax (\textcolor{violet}{\textit{gram}}), datasets] [\textit{\underline{Spelling and punctuation}} (\textcolor{violet}{\textit{spell}}), datasets [Spelling correctness (\textcolor{violet}{\textit{spcorr}}), models] [Punctuation correctness (\textcolor{violet}{\textit{puncorr}}), models] ] [Lexical choice and diversity (\textcolor{violet}{\textit{div}}), datasets] ] [Coherence (\textcolor{violet}{\textit{coh}}), domain_box [Topic relevance (\textcolor{violet}{\textit{topic}}), datasets] [Logical flow (\textcolor{violet}{\textit{logic}}), datasets] [Context consistency (\textcolor{violet}{\textit{context}}), datasets] ] [Engagingness (\textcolor{violet}{\textit{eng}}), domain_box [\textit{\underline{Content richness}} (\textcolor{violet}{\textit{contr}}), datasets [Information quantity (\textcolor{violet}{\textit{infoquant}}), models] [Information quality (\textcolor{violet}{\textit{infoqual}}), models] [Topic diversity (\textcolor{violet}{\textit{topdiv}}), models] [Topic relevance (\textcolor{violet}{\textit{toprel}}), models] ] [\textit{\underline{Emotional engagement}} (\textcolor{violet}{\textit{emo}}), datasets [Length (\textcolor{violet}{\textit{length}}), models] [Tone (\textcolor{violet}{\textit{tone}}), models] [Engagement (\textcolor{violet}{\textit{engage}}), models] ] [Feedback (\textcolor{violet}{\textit{feedback}}), datasets] [User involvement (\textcolor{violet}{\textit{userinv}}), datasets] ] [Groundedness (\textcolor{violet}{\textit{grd}}), domain_box [Factual consistency (\textcolor{violet}{\textit{factcon}}), datasets] [\textit{\underline{Factual accuracy}} (\textcolor{violet}{\textit{factacc}}), datasets [Factual correctness (\textcolor{violet}{\textit{factcorr}}), models] [Factual source (\textcolor{violet}{\textit{factsource}}), models] [Factual relevance (\textcolor{violet}{\textit{factrel}}), models] ] [Knowledge (\textcolor{violet}{\textit{knowle}}), datasets] [Consistency (\textcolor{violet}{\textit{con}}), datasets] ] ] \end{forest} 
\end{adjustbox} 
\caption{A case study for criteria decomposition on Topical-Chat. White, \textcolor{framework-blue}{blue} and \textcolor{paired-dark-orange}{orange} boxes denote decomposed criteria at \engordnumber{1}, \engordnumber{2} and \engordnumber{3} hierarchy. \underline{\textit{Underlined}} denote criteria being selected with attribution pruning.} 
\label{fig:tree} 
\end{figure*}
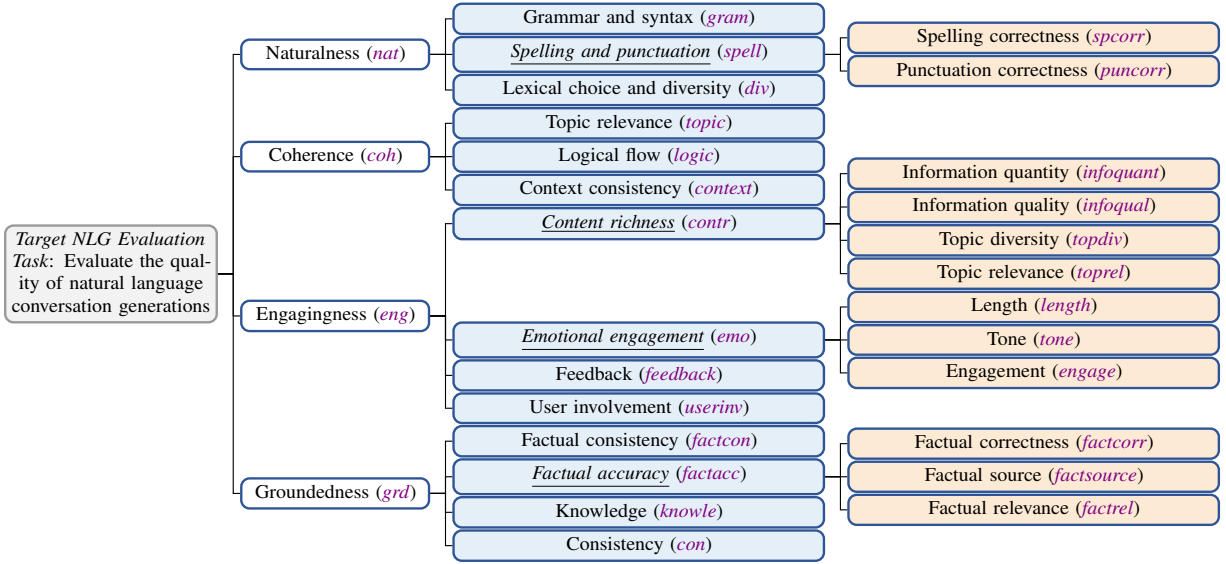

\paragraph{Ablation Study} In Table \ref{tab:aba}, we provide an ablation study on key components of \ours. We first investigate the effectiveness of hierarchical criteria decomposition, by removing layers of hierarchy in a bottom-up fashion. As illustrated in the table, the human relevance drops consistently on both correlation measurements with layers being removed, demonstrating the significance of criteria decomposition. We then replaced the human preference-guided aggregator with a numeric average on all labels, and its performance dropped significantly ($p<0.05$). These results verify that the crucial design components of \ourswb positively contribute to human alignment.

\begin{table}[t]
\center 

\tabcolsep0.07 in

\resizebox{0.49\textwidth}{!}{
\begin{tabular}{lcccccc}
\toprule
\multicolumn{1}{l}{\multirow{2}[1]{*}{\textbf{Metrics}}} & \multicolumn{2}{c}{\textbf{SummEval}}
 & \multicolumn{2}{c}{\textbf{TopicalChat}} & \multicolumn{2}{c}{\textbf{SFHOT}}  \\
 
 & $r$ & $\rho$ &  $r$ & $\rho$ & $r$ & $\rho$ \\
 
\cmidrule(lr){1-1} \cmidrule(lr){2-3} \cmidrule(lr){4-5} \cmidrule(lr){6-7}

\multicolumn{7}{c}{\cellcolor[rgb]{ .886,  .937,  .855} \textit{Iterative alignment training on $\textbf{50}\%$ of data}} \\
\addlinespace[0.3ex]
\ours\textsc{-NN} & \textbf{0.617} & \textbf{0.535} & \textbf{0.616} & \textbf{0.638} & \textbf{0.510} & \textbf{0.432} \\ 
$\quad$ w/o Layer 3 & 0.611 & 0.534 & 0.600 & 0.624 & 0.470 & 0.356 \\ 
$\quad$ w/o Layer 2,3 & 0.576 & 0.516 & 0.535 & 0.543 & 0.448 & 0.346\\ 
$\quad$ w/o Layer 1,2,3 & 0.538 & 0.513 & 0.567 & 0.590 & 0.436 & 0.364 \\ 
$\quad$ w/o Aggregator & 0.555 & 0.530 & 0.600 & 0.615 & 0.406 & 0.313 \\




\bottomrule
\end{tabular}
}
\caption{Ablations on each proposed module of \ours. We report Pearson ($r$) and Spearman ($\rho$) correlations on all NLG evaluation tasks explored in this study.}
\label{tab:aba}
\end{table}
\paragraph{Aggregator Implementation} We explore various implementations of human preference estimator in \ours. As listed in Table \ref{tab:agg-impl}, more capable aggregators like random forest or shallow NNs contribute to a better alignment in general, while a simplistic linear regression also stays on-par on most tasks, and even excels at Data-to-Text tasks.
\begin{table}[t]
\center 

\tabcolsep0.07 in

\resizebox{0.49\textwidth}{!}{
\begin{tabular}{lcccccc}
\toprule
\multicolumn{1}{l}{\multirow{2}[1]{*}{\textbf{Metrics}}} & \multicolumn{2}{c}{\textbf{SummEval}}
 & \multicolumn{2}{c}{\textbf{TopicalChat}} & \multicolumn{2}{c}{\textbf{SFHOT}}  \\
 
 & $r$ & $\rho$ &  $r$ & $\rho$ & $r$ & $\rho$ \\
 
\cmidrule(lr){1-1} \cmidrule(lr){2-3} \cmidrule(lr){4-5} \cmidrule(lr){6-7}

\multicolumn{7}{c}{\cellcolor[rgb]{ .886,  .937,  .855} \textit{Iterative 
alignment training on $\textbf{25}\%$ of data}} \\
\addlinespace[0.3ex]
\ours-LR & 0.568 & 0.521 & 0.495 & 0.519 & 0.448 & 0.390 \\ 
\ours-DT & 0.488 & 0.442 & 0.401 & 0.398 & 0.397 & 0.347 \\ 
\ours-RF & \textbf{0.607} & 0.502 & 0.589 & 0.602 & 0.413 & 0.366 \\ 
\ours-NN & 0.598 & \textbf{0.529} & \textbf{0.591} & \textbf{0.621} & \textbf{0.494} &  \textbf{0.420} \\ \midrule

\multicolumn{7}{c}{\cellcolor[rgb]{ .886,  .937,  .855} \textit{Iterative 
alignment training on $\textbf{50}\%$ of data}} \\
\addlinespace[0.3ex]
\ours-LR & 0.583 & 0.534 & 0.599 & 0.617 & \textbf{0.512} & \textbf{0.443} \\ 
\ours-DT & 0.505 & 0.430 & 0.525 & 0.549 & 0.330 & 0.274 \\ 
\ours-RF & 0.614 & 0.504 & 0.615 & 0.626 & 0.480 & 0.397 \\ 
\ours-NN & \textbf{0.617} & \textbf{0.535} & \textbf{0.616} & \textbf{0.638} & 0.510 &  0.432 \\
\bottomrule
\end{tabular}
}
\caption{Exploring \ourswb varying implementation of aggregator. We report Pearson ($r$) and Spearman ($\rho$) correlations on all NLG evaluation tasks in this study.}
\label{tab:agg-impl}
\end{table}

\section{Analysis}
\label{ch:analysis}
\subsection{Case Study}
To investigate the effect of hierarchical criteria decomposition, we present a case study on evaluating natural language conversation. In our experiments, we explore decomposing an NLG evaluation task into a maximum of 3 hierarchies (layers). As illustrated in Figure \ref{fig:tree}, the highest layer of \ourswb resembles \textit{high-level} evaluation aspects focusing on holistic evaluations, e.g. naturalness and coherence. These holistic criteria are then elaborated and supported with finer-grained decomposition at layer 2, focusing on \textit{more specific} aspects. The last layer further expands attributed significant ones to \textit{finest-grained} criteria. These results demonstrate the capability of \ourswb in generating hierarchical criteria decomposition for NLG evaluations. A complete case study is presented in Appendix \ref{app:case}.

\begin{figure*}[ht]
  \centering
  \includegraphics[width=0.93\textwidth]{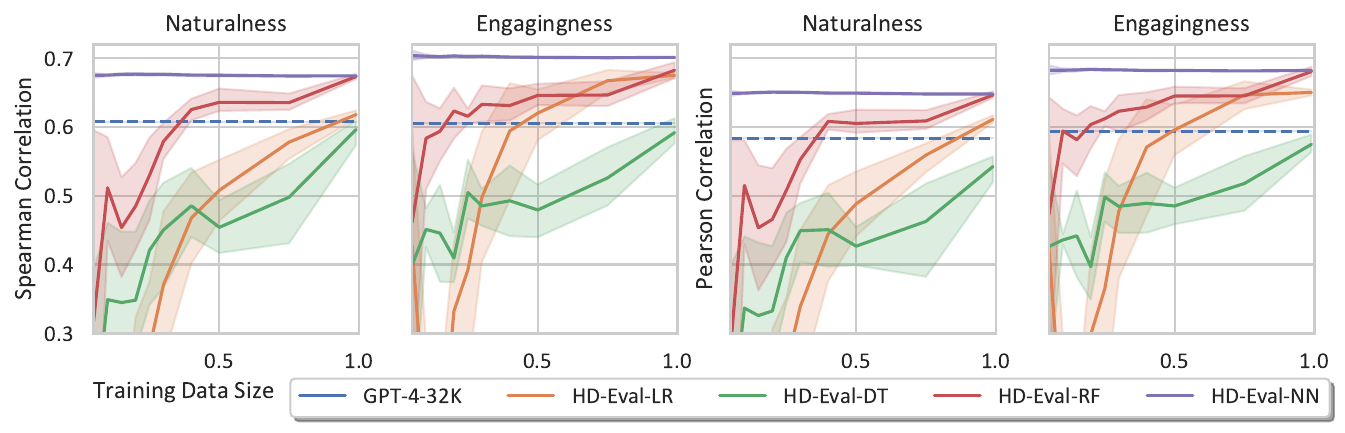}
  \caption{Performance of \ourswb under different training data counts on Topical-Chat, averaged over 5 seeds.}
  \label{fig:data_eff}
\end{figure*}

\begin{figure}[t]
  \centering
  \includegraphics[width=0.46\textwidth]{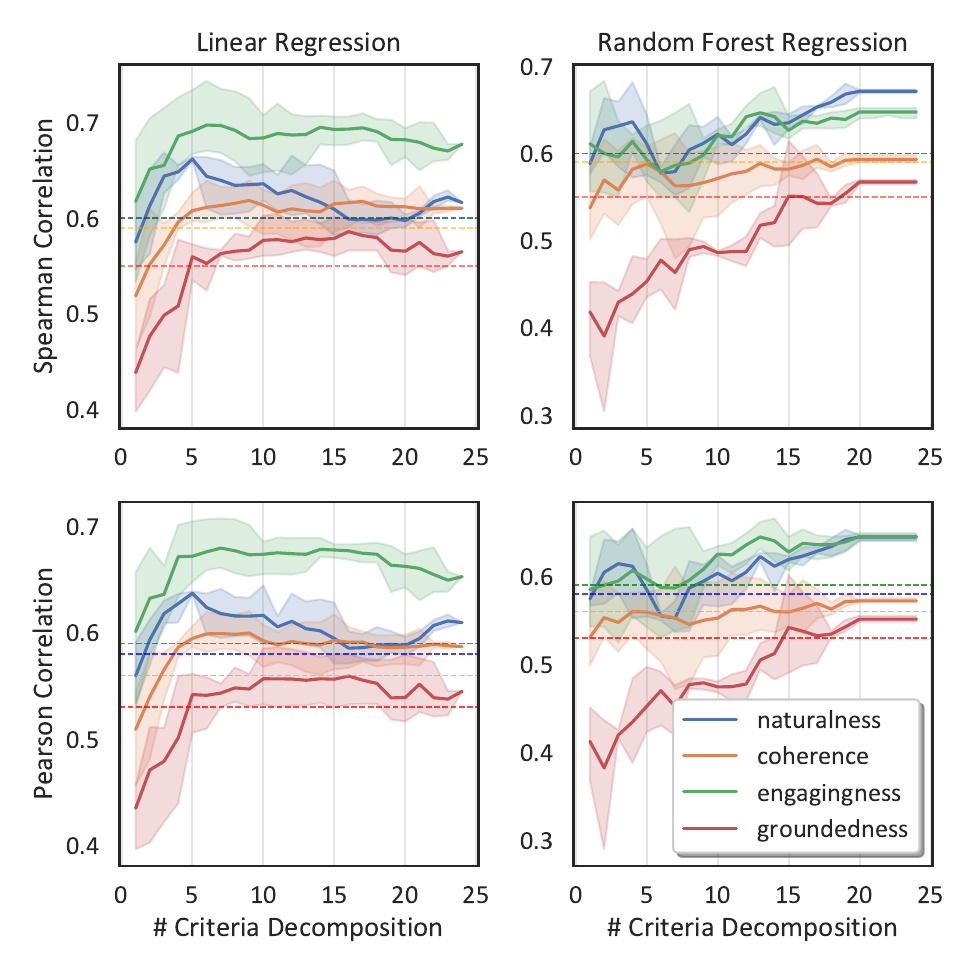}
  \caption{Criteria efficiency of \ourswb on Topical-Chat. Results are averaged over 5 random samples.}
  \label{fig:label_eff}
\end{figure}

\begin{figure}[ht]
  \centering
  \includegraphics[width=0.46\textwidth]{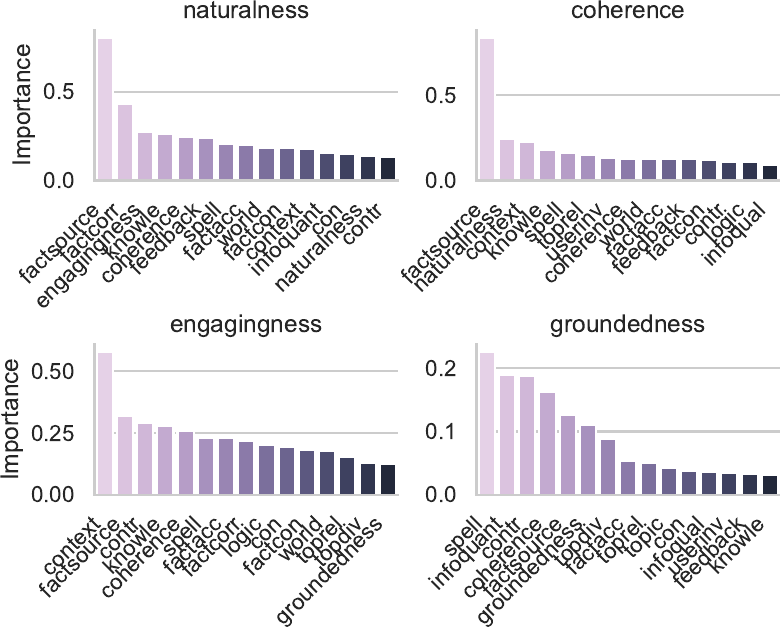}
  \caption{Explainability on preference estimation of \ours\textsc{-NN} based on permutation importance.}
  \label{fig:crit_eff}
\end{figure}

\subsection{Data Efficiency} In Section \ref{ch:results}, we demonstrate \ourswb is significant in aligning LLM-based evaluators. However, this also requires annotations from experts. To test \ourswb under different amounts of data, we sweep training data percentage from $5\%$ to full corpus. As illustrated in Figure \ref{fig:data_eff}, more data generally benefits \ourswb in improving human alignment, as it provides more evidence to infer the underlying pattern of human mindsets. A stronger regressor reduces the demand on human labels (e.g. only training on $5\%$ of data is sufficient for \ours-NN).
This intriguing feature ensures an efficient deployment and uncovers the fact that such alignment is rather \textit{superficial}, which corroborates with \citet{zhou2023lima}. Once we obtain a decomposition, the remaining efforts on addressing human preference are thereby light, since it should be \textit{shared implicitly as a `consensus'} within human experts.

\subsection{Criteria Efficiency}
While the search space of \ourswb has already been significantly reduced with attribution pruning, we investigate whether a \textit{post-pruning} could be performed on top of it. To investigate, we first sort all decomposed criteria (nodes) via significance, then progressively add them and train proxy aggregators. Results are illustrated in Figure \ref{fig:label_eff}. Generally, since more information is provided, increasing criteria counts contribute to a better alignment. However, it is also proven feasible to achieve a comparable performance by only keeping the most significant ones for better efficiency\footnote{While post-pruning greatly benefits efficiency, this does not undermine the significance of criteria decomposition, since with which we search for fine-grained candidate criteria.}.

\subsection{Explainability of \ours}
In this subsection, we discuss the explainability of the evaluation results generated with \ours. To provide a lens of interpretation, we implement human preference-guided aggregators in a lightweight, white-box fashion, providing us with possibilities in post-hoc explanations. We experiment with two attribution approaches: permutation importance \citep{altmann2010permutation} and Sharply additive explanations \citep{lundberg2017unified}.

\begin{figure}[t]
  \centering
  \includegraphics[width=0.47\textwidth]{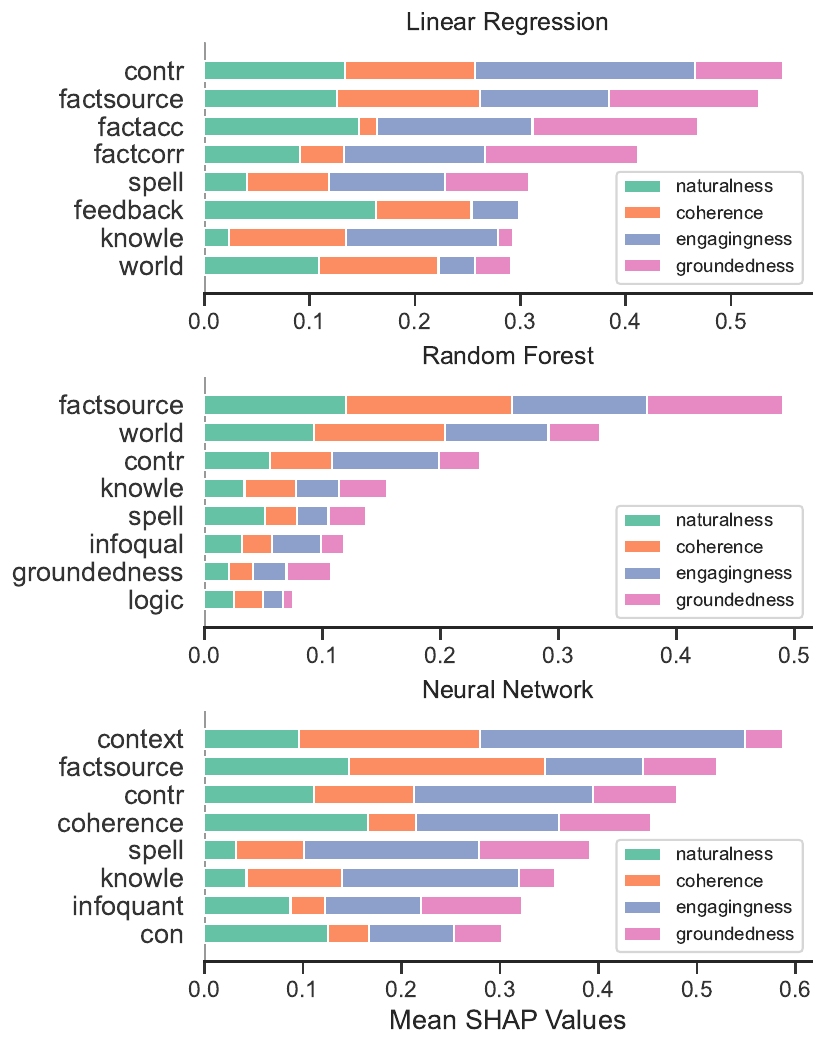}
  \caption{Explainability on human preference estimation of \ourswb based on SHAP.}
  \label{fig:crit_eff_shap}
\end{figure}

As illustrated in Figure \ref{fig:crit_eff} and \ref{fig:crit_eff_shap},\ourswb successfully assigned importance to various decomposed criteria as an estimation of human preference for different evaluation aspects, indicating the effectiveness in the human preference-guided aggregation process of \ours. 
These results also provide a lens into \textit{understanding underlying human preference} from evaluation. For instance, we mine and uncover multiple crucial \textit{key objectives} for dialogue generation, including factual correctness (\textit{factcorr}), content richness (\textit{contr}), factual source (\textit{factsource}), which are shared by all target evaluation aspects. These findings above not only improve our understanding of human preference in evaluation but also provide key grasps into \textbf{\textit{directions}} of refining candidate models (e.g., LLMs). 

\section{Related Work}
\label{ch:rel}

\paragraph{Automatic Text Evaluation}
Conventional metrics like BLEU \citep{papineni2002bleu} and ROUGE \citep{lin2004rouge} assess candidate quality by statistically comparing n-grams with a reference text, but their human alignment is criticized \citep{freitag2022stopbleu}. In contrast, embedding-based metrics, using PLM embeddings like BERT \citep{bert}, gauge similarity between candidate and reference \citep{Zhang*2020BERTScore:, zhao2019moverscore}, yet they are limited by their reliance on a similarity-based approach and the quality and diversity of references.

More recent research aims to enhance PLMs through fine-tuning on human \citep{rei2020comet} or synthetic \citep{zhong2022unieval} labels, or pretraining on domain-relevant documents \citep{yuan2021bartscore}. However, metrics in these studies either emphasize a single dimension \citep{wang2020asking, huang2020grade} or are limited in human relevance \citep{mehri2020usr, zhong2022unieval}.

\paragraph{LLM-Based Evaluators}
As LLMs gain prominence, recent research delves into the development of LLM-based evaluators. Early investigations involve initial explorations on LLMs, including prompting methods and model variants \citep{fu2023gptscore, kocmi2023gemba, wang2023chatgptgood, chen2023exploring, liu2023gpteval}. 

A subsequent line of studies aims to address extant limitations within these evaluators, with a focus on factors such as factuality \citep{min2023factscore}, interpretability \citep{lu2023error}, mitigation of position bias \citep{wang2023notfair}, and alignment to human evaluation standards \citep{liu2023calibrating}.
Another strand of works explores empowering LLM-based evaluation methodologies. This involves efforts directed at generalization to underrepresented languages \citep{hada2023large-multi}, grounding evaluations into error spans \citep{fernandes2023devil}, incorporating interactive discussions \citep{chan2023chateval}, and human collaboration \citep{li2023collaborative}.
Diverging from these approaches, we focus on the iterative alignment of LLM-based evaluators through hierarchical criteria decomposition and are the first to break down evaluation into a hierarchy of criteria at different granularity.

\section{Conclusion}
Drawing inspiration from human evaluation mindsets, we propose \ours, a novel framework that empowers LLM-based evaluators through explainable alignment. Through criteria decomposition, human preference-guided aggregation, and attribution pruning, the criteria obtained with \ourswb demonstrates a comprehensive focus on different levels of details. Extensive experiments on three NLG evaluation tasks demonstrate the effectiveness of \ours. Detailed analysis shows the efficiency and explainability of \ours, and opens up brand new perspectives in understanding preferences of human evaluations.

\newpage
\section*{Limitations} 
Below, we make an elaborate discussion about the current limitations of this work and share our perspectives on further directions.

\begin{enumerate}[1)]
  \item Currently, criteria decomposition in this work is solely done with LLMs in this work due to the lack of domain knowledge and limited resources. Ideally, \ourswb would exploit its full potential by leveraging \textit{human-in-the-loop} to assist the criteria decomposition and iterative pruning procedure. Also, it could be potentially beneficial to employ expert-written guidelines for each evaluation aspect. We leave this as a promising direction for future work. 
  \item The underlying assumption of \ourswb is that an evaluation task is \textit{decomposable}, i.e., it could be hierarchically decomposed to aspects at multiple detail levels. While this claim is natural as it follows the essence of human evaluation mindsets, it remains elusive whether we can always optimally decompose a task hierarchically, which demands future investigations and possible improvements.
  \item Limited by scope and budget, we did not perform exhaustive research on prompt engineering for LLM-based evaluators in \ours. As evidenced by multiple concurrent works, LLM-based evaluators are sensitive to prompts and would enjoy a performance uplift with carefully engineered prompts. We believe these research efforts are \textit{orthogonal} with \ours, and propose \ourswb as a methodology that is able to adapt to different prompts and leverage more advanced prompt designs in the future. 
\end{enumerate}

\section*{Ethnics Statement}
\ourswb aims to improve the evaluation of natural language generation systems by using a novel framework that aligns LLM-based evaluators with human preference. This work has the potential to benefit the research community and society by providing more reliable and transparent metrics for assessing the quality of NLG outputs. 

This work also acknowledges the possible risks and challenges associated with using LLMs for evaluation, such as the potential bias against the contents generated by different systems, the ethical and legal implications of using LLMs that may contain sensitive or harmful information, and the computational and environmental costs of training and deploying LLMs.

All language models and human annotations applied throughout this study are publicly available, and properly cited in relevant sections of this paper.




\bibliography{anthology,custom,coling}
\appendix
\section{Extended Analysis}
\label{app:anal}
\begin{figure*}[t]
  \centering
  \includegraphics[width=0.97\textwidth]{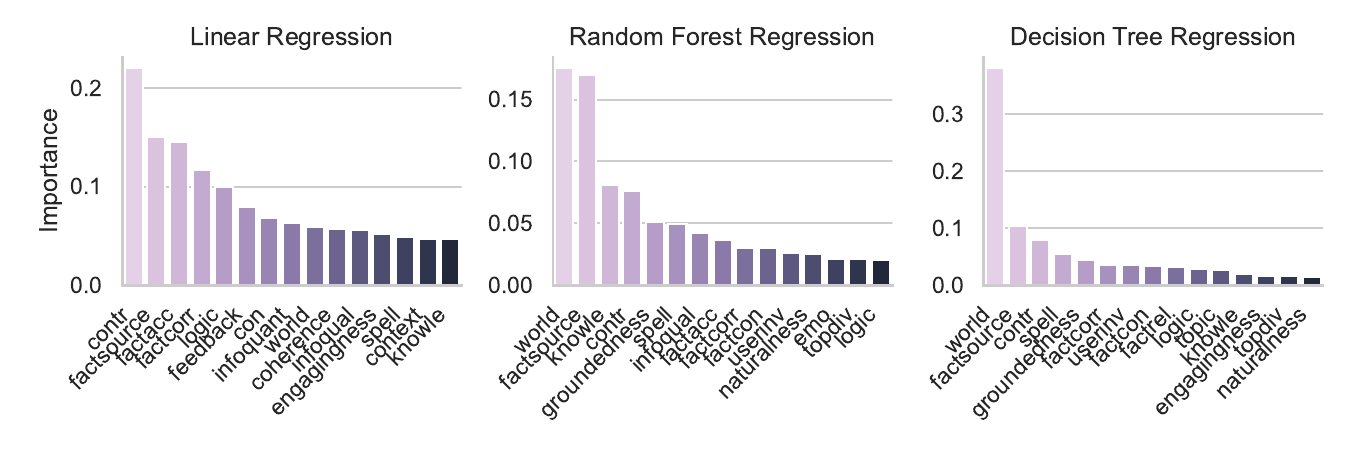}
  \caption{Explaiability on human preference estimation of \ours, based on permutation importance (LR) and weights (Tree-Based implementations), on Topical-Chat.}
  \label{fig:crit_eff_ex_p}
\end{figure*}

\begin{figure*}[t]
  \centering
  \includegraphics[width=0.97\textwidth]{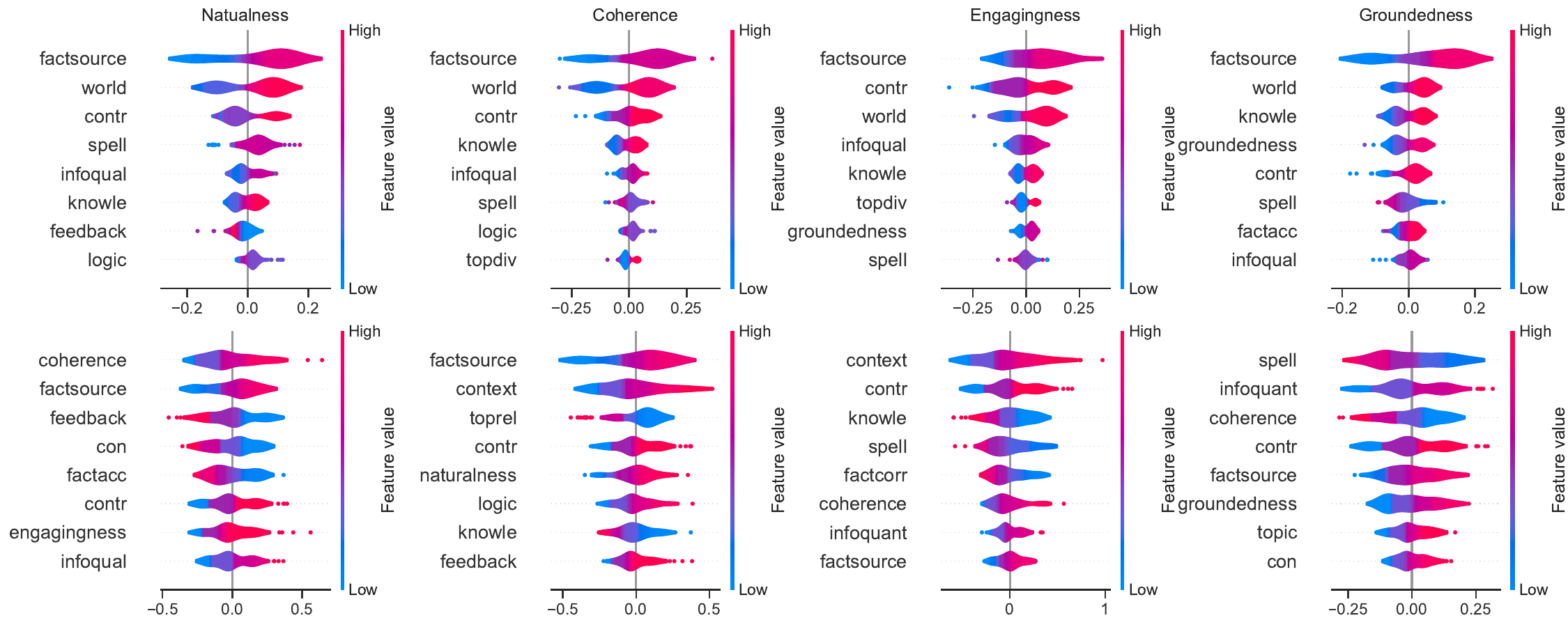}
  \caption{Explaiability on human preference estimation of \ours-RF and \ours-NN, based on shapley additive values, on Topical-Chat. A total count of 100 samples are randomly selected for attribution.}
  \label{fig:crit_eff_ex_s}
\end{figure*}

In this subsection, we provide an extended analysis of the explainability of evaluations of \ours. Results are presented in Figure \ref{fig:crit_eff_ex_p} and \ref{fig:crit_eff_ex_s}. In Figure \ref{fig:crit_eff_ex_p}, we perform permutation importance analysis on other implementations of \ourswb in addition to Figure \ref{fig:crit_eff}. In figure \ref{fig:crit_eff_ex_s}, we perform a detailed visualization of SHAP (Shapley additive explanation values) on \ours-NN and \ours-RF.

From these results, we observe that Tree-based (DT, RF) and Regression-based (LR, NN) demonstrate similar traits in assigning importance to decomposed criteria. However, our conclusion still holds that a set of underlying evaluation criteria are shared as critical contributors to all evaluation aspects, e.g. content richness (\textit{contr}) and factual source (\textit{factsource}). We believe the explainability of \ourswb provides a valuable perspective in understanding inherent preferences for human experts, which has potential on both qualifying human evaluations (e.g. estimating annotator bias) and providing detailed supporting evidence for improving NLG systems.

\section{Discussions On Smaller LLMs} 
\label{apd:llama}
Most previous research on LLM-based evaluations reveals that reference-free text quality evaluation is indeed a challenging task that demands immense pre-training knowledge and emergent capabilities of LLMs. 

Particularly, only a very few \textit{most capable} LLMs (e.g. GPT-4 \cite{OpenAI2023GPT4TR}) could be prompted as a strong evaluator, and zero-shot performances of smaller LLMs (e.g. Llama \citep{touvron2023llama} or Falcon-40B \citep{almazrouei2023falcon}) are largely undesired in following instructions on evaluation \citep{chiang-lee-2023-closer}. As studied in \citet{shen-etal-2023-large-not}, even the most capable \textsc{LLama-2-Chat-70B} correlates poorly with human evaluations, falling behind dedicated-tuned small neural evaluators \citep{zhong2022unieval}.

\begin{table}[t]
\center  

\tabcolsep0.07 in

\resizebox{0.48\textwidth}{!}{
\begin{tabular}{lcccccccc}
\toprule
\multicolumn{1}{l}{\multirow{2}[1]{*}{\textbf{Metrics}}} & \multicolumn{2}{c}{\textbf{Nat.}}
 & \multicolumn{2}{c}{\textbf{Coh.}} & \multicolumn{2}{c}{\textbf{Eng.}} & \multicolumn{2}{c}{\textbf{Grd.}} \\
 & $r$ & $\rho$ &  $r$ & $\rho$ & $r$ & $\rho$ & $r$ & $\rho$  \\
\cmidrule(lr){1-1} \cmidrule(lr){2-3} \cmidrule(lr){4-5} \cmidrule(lr){6-7} \cmidrule(lr){8-9} 
\multicolumn{9}{c}{\cellcolor[rgb]{ .886,  .937,  .855} \textit{Iterative alignment training on $\textbf{50}\%$ of data}} \\

Llama2-7B-Chat & 0.078 & 0.233 & 0.257 & 0.360 & 0.594 & 0.605 & 0.062 & 0.127 \\
\textbf{+\ours-RF} & 0.355 & 0.377 & 0.378 & 0.371 & 0.463 & 0.462 & 0.241 & 0.227 \\
+\ours-NN & 0.245 & 0.266 & 0.208 & 0.269 & 0.176 & 0.239 & 0.046 & 0.104 \\
Gain (\%) &\textcolor{teal}{\textbf{355.1}} & \textcolor{teal}{\textbf{61.8}} & \textcolor{teal}{\textbf{47.1}} & \textcolor{teal}{3.1} & \textcolor{purple}{-22.1} & \textcolor{purple}{-23.6} & \textcolor{teal}{\textbf{288.7}} & \textcolor{teal}{\textbf{78.7}} 
\\
\midrule

Llama2-13B-Chat & 0.371 & 0.378 & 0.295 & 0.302 & 0.594 & 0.605 & 0.269 & 0.296 \\
\textbf{+\ours-RF} & 0.353 & 0.375 & 0.378 & 0.383 & 0.528 & 0.524 & 0.357 & 0.362  \\
+\ours-NN & 0.391 & 0.386 & 0.255 & 0.250 & 0.364 & 0.400 & 0.165 & 0.160 \\
Gain (\%) & \textcolor{purple}{-4.9} & \textcolor{purple}{-0.8} & \textcolor{teal}{28.1} & \textcolor{teal}{26.8} & \textcolor{purple}{-11.1} & \textcolor{purple}{-13.4} & \textcolor{teal}{\textbf{32.7}} & \textcolor{teal}{22.3} \\
\midrule
\multicolumn{9}{c}{\cellcolor[rgb]{ .886,  .937,  .855} \textit{Iterative alignment training on $\textbf{80}\%$ of data}} \\

Llama2-7B-Chat & 0.018 & 0.159 & 0.209 & 0.333 & 0.602 & 0.616 & 0.105 & 0.073 \\
\textbf{+\ours-RF} & 0.420 & 0.397 & 0.495 & 0.436 & 0.469 & 0.469 & 0.245 & 0.203 \\
+\ours-NN & 0.501 & 0.450 & 0.508 & 0.442 & 0.453 & 0.412 & 0.216 & 0.219 \\
Gain (\%) & \textcolor{teal}{\textbf{2233.3}} & \textcolor{teal}{\textbf{149.7}} & \textcolor{teal}{\textbf{136.8}} & \textcolor{teal}{\textbf{30.9}} & \textcolor{purple}{-22.1} & \textcolor{purple}{-23.9} & \textcolor{teal}{\textbf{133.3}} & \textcolor{teal}{\textbf{178.1}} 
 \\
\midrule

Llama2-13B-Chat & 0.484 & 0.471 & 0.336 & 0.397 & 0.602 & 0.616 & 0.232 & 0.248  \\
+\ours-RF & 0.412 & 0.411 & 0.454 & 0.472 & 0.455 & 0.462 & 0.327 & 0.334 \\
\textbf{+\ours-NN} & 0.550 & 0.529 & 0.470 & 0.505 & 0.523 & 0.543 & 0.256 & 0.244  \\
Gain (\%) & \textcolor{teal}{13.6} & \textcolor{teal}{12.3} & \textcolor{teal}{\textbf{39.9}} & \textcolor{teal}{27.2} & \textcolor{purple}{-13.1} & \textcolor{purple}{-11.9} & \textcolor{teal}{10.3} & \textcolor{purple}{-1.6} 
  \\

\bottomrule
\end{tabular}
}
\caption{Exploring \ourswb on Topical-Chat with smaller LLMs. We report Pearson ($r$) and Spearman ($\rho$) correlations. Gain (\%) denote the relative performance gain from best overall performing system (marked in \textbf{bold}). We highlight relative performance gains over 30\% through \ourswb with \textcolor{teal}{\textbf{bold}}.}
\label{tab:aba-small}
\end{table}

\begin{table}[t]
\center  

\tabcolsep0.07 in

\resizebox{0.48\textwidth}{!}{
\begin{tabular}{lcccccccc}
\toprule
\multicolumn{1}{l}{\multirow{2}[1]{*}{\textbf{Metrics}}} & \multicolumn{2}{c}{\textbf{Coh.}}
 & \multicolumn{2}{c}{\textbf{Con.}} & \multicolumn{2}{c}{\textbf{Flu.}} & \multicolumn{2}{c}{\textbf{Rel.}} \\
 & $r$ & $\rho$ &  $r$ & $\rho$ & $r$ & $\rho$ & $r$ & $\rho$  \\
\cmidrule(lr){1-1} \cmidrule(lr){2-3} \cmidrule(lr){4-5} \cmidrule(lr){6-7} \cmidrule(lr){8-9} 

\multicolumn{9}{c}{\cellcolor[rgb]{ .886,  .937,  .855} \textit{Iterative alignment training on $\textbf{20}\%$ of data}} \\

Llama2-7B-Chat & 0.097 & 0.096 & 0.008 & 0.005 & 0.034 & 0.024 & 0.134 & 0.130 \\
+\ours-RF & 0.054 & 0.053 & 0.058 & 0.049 & 0.025 & 0.010 & 0.151 & 0.150 \\
\textbf{+\ours-NN} & 0.138 & 0.132 & 0.130 & 0.061 & 0.111 & 0.071 & 0.130 & 0.123 \\
Gain (\%) & \textcolor{teal}{\textbf{42.3}} & \textcolor{teal}{\textbf{37.5}} & \textcolor{teal}{\textbf{1525.0}} & \textcolor{teal}{\textbf{1120.0}} & \textcolor{teal}{\textbf{226.5}} & \textcolor{teal}{\textbf{195.8}} & \textcolor{purple}{-3.0} & \textcolor{purple}{-5.4} 
 
 \\
\midrule

Llama2-13B-Chat & 0.268 & 0.246 & 0.134 & 0.114 & 0.138 & 0.124 & 0.132 & 0.118  \\
\textbf{+\ours-RF} & 0.267 & 0.227 & 0.244 & 0.130 & 0.197 & 0.137 & 0.278 & 0.212 \\
+\ours-NN & 0.299 & 0.277 & 0.141 & 0.100 & 0.160 & 0.098 & 0.250 & 0.220 \\
Gain (\%) & \textcolor{purple}{-0.4} & \textcolor{purple}{-7.7} & \textcolor{teal}{\textbf{82.1}} & \textcolor{teal}{14.0} & \textcolor{teal}{\textbf{42.8}} & \textcolor{teal}{10.5} & \textcolor{teal}{\textbf{110.6}} & \textcolor{teal}{\textbf{79.7}} 
  \\
\midrule

Llama2-70B-Chat & 0.392 & 0.383 & 0.277 & 0.232 & 0.248 & 0.217 & 0.304 & 0.254 \\
+\ours-RF & 0.408 & 0.367 & 0.249 & 0.214 & 0.233 & 0.164 & 0.409 & 0.370   \\
\textbf{+\ours-NN} & 0.454 & 0.418 & 0.306 & 0.206 & 0.311 & 0.214 & 0.451 & 0.421 \\
Gain (\%) & \textcolor{teal}{15.8} & \textcolor{teal}{9.1} & \textcolor{teal}{10.5} & \textcolor{purple}{-11.2} & \textcolor{teal}{25.4} & \textcolor{purple}{-1.4} & \textcolor{teal}{\textbf{48.4}} & \textcolor{teal}{\textbf{65.7}} 
 \\

\midrule
 
\multicolumn{9}{c}{\cellcolor[rgb]{ .886,  .937,  .855} \textit{Iterative alignment training on $\textbf{50}\%$ of data}} \\

Llama2-7B-Chat & 0.064 & 0.064 & 0.010 & 0.017 & 0.001 & 0.032 & 0.127 & 0.133 \\
\textbf{+\ours-RF} & 0.118 & 0.124 & 0.131 & 0.182 & 0.062 & 0.055 & 0.216 & 0.200 \\
+\ours-NN & 0.103 & 0.109 & 0.169 & 0.100 & 0.085 & 0.081 & 0.147 & 0.140 \\
Gain (\%) & \textcolor{teal}{\textbf{84.4}} & \textcolor{teal}{\textbf{93.8}} & \textcolor{teal}{\textbf{1210.0}} & \textcolor{teal}{\textbf{970.6}} & \textcolor{teal}{\textbf{6100.0}} & \textcolor{teal}{\textbf{71.9}} & \textcolor{teal}{\textbf{70.1}} & \textcolor{teal}{\textbf{50.4}} 
 
\\
\midrule

Llama2-13B-Chat & 0.235 & 0.219 & 0.119 & 0.109 & 0.142 & 0.110 & 0.148 & 0.148 \\
\textbf{+\ours-RF} & 0.296 & 0.230 & 0.272 & 0.140 & 0.181 & 0.100 & 0.332 & 0.281  \\
+\ours-NN & 0.282 & 0.258 & 0.214 & 0.146 & 0.158 & 0.064 & 0.263 & 0.252 \\
Gain (\%) & \textcolor{teal}{26.0} & \textcolor{teal}{5.0} & \textcolor{teal}{\textbf{128.6}} & \textcolor{teal}{28.4} & \textcolor{teal}{27.5} & \textcolor{purple}{-9.1} & \textcolor{teal}{\textbf{124.3}} & \textcolor{teal}{\textbf{89.9}} 

 \\
\midrule

Llama2-70B-Chat & 0.367 & 0.360 & 0.253 & 0.225 & 0.255 & 0.199 & 0.268 & 0.234 \\
+\ours-RF & 0.392 & 0.372 & 0.364 & 0.278 & 0.284 & 0.214 & 0.386 & 0.348  \\
\textbf{+\ours-NN} & 0.418 & 0.383 & 0.381 & 0.286 & 0.347 & 0.210 & 0.457 & 0.432 \\
Gain (\%) & \textcolor{teal}{13.9} & \textcolor{teal}{6.4} & \textcolor{teal}{\textbf{50.6}} & \textcolor{teal}{27.1} & \textcolor{teal}{\textbf{36.1}} & \textcolor{teal}{5.5} & \textcolor{teal}{\textbf{70.5}} & \textcolor{teal}{\textbf{84.6}} 
\\

\bottomrule
\end{tabular}
}
\caption{Exploring \ourswb on SummEval with smaller LLMs. We report Pearson ($r$) and Spearman ($\rho$) correlations. Gain (\%) denote the relative performance gain from best overall performing system (marked in \textbf{bold}). We highlight relative performance gains over 30\% through \ourswb with \textcolor{teal}{\textbf{bold}}.}
\label{tab:aba-small-sme}

\end{table}

To exploit the full potential of smaller language models in zero-shot evaluation, we explore empowering them with \ours. We experimented with \textsc{LLama2-Chat-7B} and \textsc{LLama2-Chat-13B}\footnote{We kept everything identical to our main experiments - same data splits, same aggregator and decomposition setting, and permutation importance for attribution pruning, except we prompt Llama for evaluation scores to each sub-criteria.}. \citep{touvron2023llama}, and results\footnote{In these tables, we mark the relative gains from the best \textit{overall} performing implementation, which may not always correspond to the best performer for a specific \textit{aspect}. We aim to present an overall effect of \ourswb on Llama models.} are illustrated in Table \ref{tab:aba-small} and \ref{tab:aba-small-sme}. On Topical-Chat, aligned with \ours, the human alignment of 7B-sized models substantially improved, achieving a 30\% or even more than 100\% improvement in evaluating the naturalness, coherence, and groundedness of conversations. Different from GPT-4, the engagingness did not obtain performance gains from hierarchical decomposition. We conjecture this phenomenon \textit{still}, roots back into poorer instruction following the capability of smaller models, where they fail to understand finer-grained, detailed evaluation aspects, as they may receive less prior knowledge in these fields. 

Similarly, \ourswb also empowers the human alignment in the evaluation of summarization quality, achieving significant gains for all 7B, 13B, and 70B variants, highlighting the universal applicability of \ours, especially when existing prompting-based methods all fall short on smaller models due to their weaker instruction following capability \citep{chiang-lee-2023-closer, shen-etal-2023-large-not}. 

Despite the gains, it is noteworthy to point out that these smaller LMs are not strong zero-shot evaluators so far. We believe a specialized and dedicated tuning \citep{gekhman2023trueteacher} on instruction following in evaluation would be a promising aid and would pursue in future endeavors.

\section{Configuration Details}
\subsection{Algorithmic Formulation}
\begin{algorithm*}[htb] 
    \caption{Iterative Alignment Training of \ours}\label{alg:mainalgo}
    \label{algo:overall}
    \begin{algorithmic}
        \STATE {\bfseries Require:} Large language model $LLM$, development set $D$, human labels $S \in \mathbb{R}^{|D|\cdot p}$, aggregator $f_\theta (\cdot)$, saliency function $g(\cdot)$, maximum hierarchical decomposition layer $L$, decomposition prompt template $T_d$, evaluation prompt template $T_e$, max decomposition child count $k$ (for any arbitrary criteria).
        \STATE {\bfseries Initialize:} An empty $A: \{A_1, ..., A_L\}$ for storing fine-grained evaluation results at each hierarchy.
    \end{algorithmic}
    \begin{algorithmic}[1]
        \FOR{iteration $j$ in $L$} 
            \STATE Initialize $C_j$ as an empty set
            \FOR{$c$ in $C_D^{j}$} 
                \STATE \texttt{\textcolor{darkblue}{// Criteria decomposition}}
                \STATE Obtain its decomposition as $LLM(T_d, c)$ and add to $C_j$
            \ENDFOR
        \FOR{sample $d$ in $D$} 
            \FOR{criteria $c_i$ in $C_j$} 
                \STATE \texttt{\textcolor{darkblue}{// Fine-grained evaluation}}
                \STATE Obtain evaluation scores $a^{j,i}_d \in \mathbb{R}$ with hierarchy-aware evaluation prompt $T_e$ and $LLM$
                \STATE Append the results $a^{j,i}_d$ to cache $A_d$
            \ENDFOR
        \ENDFOR
        \STATE \texttt{\textcolor{darkblue}{// Human preference-guided aggregation}}
        \STATE Train proxy aggregator $f_j: \mathbb{R}^{|\cup_{r<j}C_r|} \to \mathbb{R}^p$ over $A$ and target $S$
        \STATE \texttt{\textcolor{darkblue}{// Attribution pruning}}
        \STATE Identify significant criteria in $C_j$ to decompose at the next later: $C_D^{j+1} = \mathrm{argtop}k_{c \in \mathcal{C}_j} \left[g\left(f_j(c)\right)\right]$.
        \ENDFOR
    \end{algorithmic}
    \begin{algorithmic}
        \STATE {\bfseries Return:} Hierarchical criteria decomposition $\{C_1, ..., C_L\}$, Finalized aggregator $f_L$
    \end{algorithmic}
\end{algorithm*}
For a concise understanding of \ours, we provide a formal algorithmic description in Algorithm \ref{alg:mainalgo}.

\subsection{Configurations}
\label{app:impl}
For hierarchical criteria decomposition, we consider a maximum of 3 layers across this study. Details on the decomposition process are listed below.
\begin{enumerate}[1)]
  \itemsep0em
  \item For the first layer, we adopt reference decomposition (multiple evaluation aspects) from human experts in the labeled data we apply.
  \item For the second layer, we expand all nodes in layer 1, each to a maximum of 4 child. This is based on the assumption that the reference evaluation aspects designated by human experts are significant and demand further in-depth deliberate evaluation.
  \item For the third layer, we apply attribution pruning as elaborated in the paper to select nodes (criteria) to further decompose.
\end{enumerate}

\subsection{Implementation}
For GPT-4 in \ours, we sample with Temperature of $0.0$ and Top-P of $1.0$, returning a maximum of $32$ tokens. Hierarchical criteria decomposition is performed with the Creative mode of Microsoft Bing Chat\footnote{\url{bing.com/chat}}, which is also powered by GPT-4.

All aggregators are implemented with the scikit-learn \citep{pedregosa2011scikit} library. For DT and RF, we apply their default built-in parameters. 
For NN, we adopt a 3-layer shallow MLP architecture, with ReLU activation. 
Aggregators are trained to regress all decomposed criteria, to fit on a set of human-annotated evaluations as $f_\theta : \mathbb{R}^{m} \to \mathbb{R}^{n}$, where $n$ denote human annotation count for a sample, and $m = \sum_{i=1}^{L} |\mathcal{C}_i|$ equals to the total count of decomposed criteria\footnote{A separate aggregator is trained for evaluating groundedness of Topical-Chat, as it has different evaluation protocols and ranges from others.}.

\subsection{Licences}
All large language models and human annotations applied throughout this study are publicly available, and properly cited in relevant sections of this paper. We acknowledge their contribution to advancing NLG research, and enlist the open-source licenses for artifacts applied in this study below:
\begin{enumerate}[1)]
  \itemsep0em
  \item LLama-2\footnote{\url{https://huggingface.co/meta-llama/Llama-2-7b-chat-hf}} models are licensed from Meta\footnote{\url{https://ai.meta.com/resources/models-and-libraries/llama-downloads/}}.
  \item SummEval\footnote{\url{https://github.com/Yale-LILY/SummEval}} is licensed under MIT.
  \item Topical-Chat\footnote{\url{https://github.com/alexa/Topical-Chat}} is licensed under Apache-2.0.
  \item SFHOT, SFRES are licensed under MIT.
\end{enumerate}

\section{Case Study on Ranking}
In this section, we present a case study on leveraging \ourswb for ranking given multiple NLG candidates. We select the SummEval \cite{fabbri2021summeval} benchmark, as it has multiple summaries for a document from different NLG systems, which suits well for ranking them w.r.t quality. We primarily compare \ourswb with GPT-4 based evaluation \cite{liu2023gpteval}.

We first calculate the exact match in ranking order on all samples (which is a very strict standard compared to Spearman rank correlation), and the accuracy of \ourswb is $36.7\%$, significantly higher than $24.8\%$ of GPT-4 Eval. Performance gains can be sourced into multiple design improvements in \ours: 1) The hierarchical decomposition captures fine-grained multi-aspect details of candidate samples, being more comprehensive;
2) The aggregator improves the alignment to human judgements;
And 3) more importantly, we provide a continuous score as output, rather than discrete judgements from prompting, which excels at distinguishing candidates of similar quality.

\begin{table*}[t]
\center 
\tabcolsep0.1 in

\resizebox{1\textwidth}{!}{
\begin{tabular}{p{0.96\textwidth}cccc}
\toprule
\multicolumn{1}{l}{\multirow{2}[1]{*}{\textbf{Generated Summary to a News Article}}} & \multicolumn{3}{c}{\textbf{Coherence}} \\
 & Human & GPT-4 & \textbf{Ours}  \\ \cmidrule(lr){1-1} \cmidrule(lr){2-4}

Paul merson has restarted his row with burnley on sunday . Townsend was brought on in the 83rd minute for tottenham . Andros townsend scores england 's equaliser in their 1-1 friendly draw . Townsend hit a stunning equaliser for england against italy . & 2.33 & 3.0 & 2.93 \\ \midrule

Paul merson has restarted his row with andros townsend after the tottenham midfielder was brought on with only seven minutes remaining in his team 's 0-0 draw with burnley on sunday . ' Paul merson had another dig at andros townsend after his appearance for tottenham against burnley . Townsend was brought on in the 83rd minute for tottenham as they drew 0-0 against burnley . & 1.67 & 2.0 & 2.67 \\ \midrule

Tottenham drew 0-0 with Burnley at Turf Moor on Sunday . Andros Townsend was brought on in the 83rd minute for Tottenham . Paul Merson criticised Townsend 's call-up to the England squad last week . Townsend hit back at Merson on Twitter after scoring for England against Italy . & 4.00 & 2.0 & 3.29 \\ \midrule

Paul merson has restarted his row with andros townsend . The tottenham midfielder was brought on with only seven minutes remaining in his team 's 0-0 draw with burnley . Townsend was brought on in the 83rd minute for tottenham as they drew 0-0 with burnley . & 3.33 & 3.0 & 3.34 \\ \midrule

Paul merson has restarted his row with andros townsend after the tottenham midfielder was brought on with only seven minutes remaining in his team 's 0-0 draw with burnley . Merson initially angered townsend for writing in his sky sports column that ` if andros townsend can get in ( the england team ) then it opens it up to anybody . ' Paul merson had another dig at andros townsend after his appearance for tottenham against burnley . & 2.67 & 2.0 & 3.30 \\ \midrule

Paul merson has restarted his row with andros townsend after the tottenham midfielder was brought on with only seven minutes remaining in his team 's 0-0 draw with burnley on sunday . Townsend was brought on in the 83rd minute for tottenham as they drew 0-0 against burnley . Townsend hit back at merson on twitter after scoring for england against italy . & 3.33 & 3.0 & 3.89 \\ \midrule

Paul merson has restarted his row with andros townsend after the tottenham midfielder was brought on with only seven minutes remaining in his team 's 0 - 0 draw with burnley on sunday . \# rubberdub \# 7minutes , ' merson put on twitter . Merson initially angered townsend for writing in his sky sports column that ' if andros townsend can get in ( the england team ) then it opens it up to anybody . & 1.00 & 2.0 & 2.10 \\ \midrule

\multicolumn{4}{c}{\textbf{Ranking:} $\quad$ \textbf{Human} (4, 5, 0, 1, 3, 1, 6) $\quad$ \textbf{GPT-4} (0, 3, 3, 0, 3, 0, 3) $\quad$ \textbf{\ourswb} (4, 5, 3, 1, 2, 0, 6) } \\

\bottomrule
\end{tabular}
}

\caption{Case study on evaluating the coherence of summary (the corresponding article is omitted due to space)}
\label{tab:case-ranking}
\end{table*}

\begin{table*}[th]
\center
\resizebox{0.95\textwidth}{!}{
\begin{tabular}{lcccccccc}
\toprule
& COH-$r$ & COH-$\rho$ & CON-$r$ & CON-$\rho$ & FLU-$r$ & FLU-$\rho$ & REL-$r$ & REL-$\rho$ \\
\midrule
\ourswb (Ours) - Average of Human & 0.668 & 0.657 & 0.604 & 0.457 & 0.580 & 0.435 & 0.619 & 0.599 \\
\midrule
Expert1-Expert2 & 0.737 & 0.725 & 0.891 & 0.750 & 0.711 & 0.569 & 0.621 & 0.554 \\
Expert1-Expert3 & 0.601 & 0.614 & 0.904 & 0.806 & 0.727 & 0.601 & 0.490 & 0.460 \\
Expert2-Expert3 & 0.597 & 0.605 & 0.945 & 0.825 & 0.722 & 0.570 & 0.501 & 0.473 \\
\midrule
Average Human-Human corr. & 0.645 & 0.648 & 0.913 & 0.794 & 0.720 & 0.580 & 0.537 & 0.496 \\
Average MTurk-Expert corr. & 0.003 & 0.009 & -0.005 & -0.025 & 0.044 & 0.019 & 0.065 & 0.090 \\
\bottomrule
\end{tabular}
}
\caption{Expert-Expert, Expert-Human (MTurk) correlation performance on SummEval}
\label{tab:human}
\end{table*}

Furthermore, we present a case study on evaluating coherence of summary. As illustrated Table \ref{tab:case-ranking}, \textsc{GPT-4} is limited by ineffectiveness in distinguishing summary of similar quality, limited by the discrete output from prompting, thus performs poorly in ranking. However, with human preference guided aggregation, \ourswb produces continuous evaluation scores, which largely improves the ranking.

\section{Comparison to Human Evaluation}
In this section, we discuss the performance ceiling of automatic evaluation by studying the human performance in SummEval, which includes 3 annotations from human experts (representing human performance ceiling) and 5 annotation from Amazon MTurk Crowd-sourcing (representing average human performance).

As illustrated in Table \ref{tab:human}, for human experts there are some discrepancies on the judgements of coherence and relevance, where \ourswb demonstrates similar performance, while their judgements on consistency are mostly concordant. Noteworthy, the average human performance (i.e., ratings from crowd-sourcers on Amazon MTurk) compared to experts is very poor, as no correlation is shown between MTurk evaluation and expert evaluations. The high thresholds for qualified human evaluation further highlights the significance of \ourswb as a promising alternative.

\section{Discussions on Concurrent works}
We discuss and highlight the improvements of our work over a concurrent work on decomposition \cite{saha2023branch}:
\begin{enumerate}[1)]
  \itemsep0em
  \item \textit{Multi-granularity hierarchical decomposition.} \citet{saha2023branch} only decomposes a task into a single layer, while we propose a more comprehensive hierarchical decomposition to capture different levels of evaluation. Our ablations (Table \ref{tab:aba}) also show its superiority beyond single-layer decomposition.
  \item \textit{Introduction of Attribution pruning}, where we objectively select and dynamically refine the decomposition, reducing the noise of criteria decomposition.
  \item \textit{Explainable aggregation.} \citet{saha2023branch} feeds all results as a prompt to the LLM to obtain a final verdict. However, this does not address human preference and is also limited by the LLM’s bias (due to LLM's preference in how to aggregate these results). In contrast, we apply white box aggregators that could be better post-hoc explained and controlled (Chapter \ref{ch:analysis}).
\end{enumerate}

\section{Listing of Prompts}
\label{app:prompts}
\subsection{Criteria Decomposition}
During the Hierarchical Criteria Decomposition procedure in \ours, we decompose criteria into finer-grained ones by jointly drafting the finer-grained criteria and their definitions with LLMs. 
An example prompt template and use case on SummEval is illustrated in Figure \ref{fig:prompt-decmp}. Note that the prompt provided here is an example, and one may freely adapt other prompting designs and methods, as long as it accomplishes reasonable decomposition. 

\subsection{Hierarchy-Aware Evaluation}
Below, we provide a complete example of the evaluation prompt templates applied for LLMs across this study, in Figure \ref{fig:prompt-chat}, \ref{fig:prompt-summ}, and \ref{fig:prompt-d2t}. As illustrated in these figures, to preserve the hierarchical information, we prompt LLMs with both the parent criteria as well as the child criteria, while detailing the child criteria with a detailed definition.

\section{Case Study on Criteria Decomposition}
\label{app:case}
In this section, we present a complete case study on the criteria decomposition process of \ours. Specifically, we provide examples of all evaluation domains in this study, as illustrated in Table \ref{tab:decmp-case-1}, \ref{tab:decmp-case-2} and \ref{tab:decmp-case-3}. As demonstrated in these tables, we observe \ourswb is capable of hierarchically decomposing evaluation criteria into finer-grained ones and capable of generating a definition alongside to further elaborate it.

\begin{figure*}[t]
\centering
\begin{tcolorbox}[width=1\textwidth, fontupper=\small, colback=blue!2, boxrule=0.9pt] 
\textbf{A) Generic template for Hierarchical Criteria Decomposition} \\\\
I would like to perform automatic evaluation on quality of \textcolor{purple}{[Evaluation Task]}. \\

\textcolor{blue}{[Backgrounds and Definitions of Evaluation Task]}. \\

I would like to to evaluate \textcolor{purple}{[List of Criteria to Decompose]}. \\

Please give me around \textcolor{purple}{[Desired Child Count]} fine-grained evaluation critics to evaluate them. I want to obtain a final comprehensive evaluation based on an overall aggregation on fine-grained metrics. With the fine-grained metrics, I can better dispatch the evaluation task to different workers and make a better overall efficiency and accuracy. \\

\textbf{B) An example use case for SummEval}\\\\
I would like to perform  automatic evaluation on quality of \textcolor{purple}{text summarization}. \\

\textcolor{blue}{A text summarization is a shorter passage that encompasses the key details of original article but much shorter. }\\

I would like to to evaluate its \textcolor{purple}{coherence, consistency, fluency, and relevance}. \\

Please give me around \textcolor{purple}{10-15} fine-grained evaluation critics to evaluate them. I want to obtain a final comprehensive evaluation based on an overall aggregation on fine-grained metrics. With the fine-grained metrics, I can better dispatch the evaluation task to different workers and make a better overall efficiency and accuracy.
\end{tcolorbox}
\caption{Prompt for Hierarchical Criteria Decomposition in \ours. We include a generic template for criteria decomposition, as well as an actual example for SummEval.}
\label{fig:prompt-decmp}
\end{figure*}
\begin{table*}[ht]
\centering
\footnotesize
\resizebox{0.98\textwidth}{!}{
\begin{tabular}{cp{0.96\textwidth}}
\toprule
Criteria & \textbf{Criteria Decomposition and Definition} \\ 
\midrule
\multicolumn{2}{c}{
\cellcolor[rgb]{ .915,  .915,  .915} \textit{\textbf{Layer 2} Decomposition}} \\
\textit{gram}  &  Grammar and syntax: The response should follow the rules of grammar and syntax, without any ungrammatical or awkward constructions. \\
\textit{spell}  &  Spelling and punctuation: The response should have correct spelling and punctuation, without any typos or errors. \\
\textit{div}  &  Lexical choice and diversity: The response should use appropriate and varied words, without any repetition or misuse of vocabulary. \\
\textit{topic}  &  Topic relevance: The response should be relevant to the topic of the dialogue. \\
\textit{logic}  &  Logical flow: The response should have a logical flow of ideas, without any abrupt changes in topic or logic. \\
\textit{context}  &  Context consistency: The response should be consistent with the context of the dialogue. \\
\textit{contr}  &  Content richness: The response should provide rich and useful content, without any generic or vague statements. \\
\textit{emo}  &  Emotional engagement: The response should be emotionally engaging, without any emotionally inappropriate statements. \\
\textit{feedback}  &  Feedback: The responsiveness and attentiveness of the dialogues to the user’s input and feedback. \\
\textit{userinv}  &  User involvement: The response should involve the user in the dialogue, without any one-sided or self-centered statements. \\
\textit{factcon}  &  Factual consistency: The response should be factually consistent, without any factual errors or contradictions. \\
\textit{factacc}  &  Factual accuracy: The response should be factually accurate, without any without any false or misleading information. \\
\textit{knowle}  &  Knowledge: The plausibility and reasonableness of the knowledge in the dialogues. \\
\textit{con}  &  Consistency: The response should be consistent with the user’s input and feedback. \\
\textit{world}  &  World knowledge: The response should demonstrate knowledge of the world, without any statements that are inconsistent with the real world. \\
\midrule
\multicolumn{2}{c}{
\cellcolor[rgb]{ .915,  .915,  .915} \textit{\textbf{Layer 3} Decomposition}}  \\
\textit{infoquant}  &  Information quantity:  The response shoulf convey adequate information, without being too brief or too verbose. \\
\textit{infoqual}  &  Information quality: The response should provide accurate, reliable, and credible content, and supported by evidence or sources. \\
\textit{topdiv}  &  Topic diversity: The response should adequate cover topics of dialogue history, without any repetition or narrow focus. \\
\textit{toprel}  &  Topic relevance: The response should match the user’s query and dialogue context, without any inconsistent or off-topic statements. \\
\textit{spcorr}  &  Spelling correctness: The response should have correct spelling, without any typos or errors. \\
\textit{puncorr}  &  Punctuation correctness: The response should have correct punctuation, without any missing or incorrect punctuation. \\
\textit{factcorr}  &  Factual correctness: The response should be factually correct, without any false or misleading information. \\
\textit{factsource}  &  Factual source: The response should be supported by reliable and credible evidence or sources, without any unsupported information or hallucinations. \\
\textit{factrel}  &  Factual relevance: The response should be relevant to the user’s query and dialogue context, being helpful instead of distracting \\
\textit{length}  &  Length: The response should be of adequate length, without being too brief or too verbose. \\
\textit{tone}  &  Tone: The response should be polite, friendly, and empathetic, without any rude or offensive statements. \\
\textit{engage}  &  Engagement: The response should be engaging and encourage further interaction, without any generic or vague statements. \\

\bottomrule
\end{tabular}}
\caption{A complete case study for criteria decomposition on Topical-Chat.}
\label{tab:decmp-case-1}
\end{table*}
\begin{table*}[ht]
\centering
\footnotesize
\resizebox{0.98\textwidth}{!}{
\begin{tabular}{cp{0.96\textwidth}}
\toprule
Criteria & \textbf{Criteria Decomposition and Definition} \\ 
\midrule
\multicolumn{2}{c}{
\cellcolor[rgb]{ .915,  .915,  .915} \textit{\textbf{Layer 2} Decomposition}} \\
\textit{ord}  &  Sentence ordering: how well the sentences in the summary follow a natural and logical order. \\
\textit{struc}  &  Discourse structure: how well the summary uses discourse markers (such as however, therefore, etc.) to indicate the relations between sentences. \\
\textit{focus}  &  Topic focus: how well the summary maintains a consistent topic throughout. \\
\textit{fact}  &  Factuality: how well the summary preserves the factual information from the original article without introducing errors or distortions. \\
\textit{entcon}  &  Entity consistency: how well the summary uses consistent names and references for entities (such as people, places, etc.) across sentences. \\
\textit{tmpcon}  &  Temporal consistency: how well the summary uses consistent tense and aspect for events across sentences. \\
\textit{gram}  &  Grammar: how well the summary use appropriate vocabulary, syntax and punctuation, and convey the main information and meaning of the article, without grammatical errors. \\
\textit{engage}  &  Engagingness: how well the summary is engaging and interesting to read. \\
\textit{read}  &  Readability: how well the summary is easy to read and understand by humans, without errors or awkward expressions. \\
\textit{cov}  &  Coverage: how well the summary includes all or most of the important information from the original article. \\
\textit{red}  &  Redundancy: how well the summary avoids repeating information that has already been mentioned or implied. \\
\textit{nov}  &  Novelty: how well the summary introduces new information that is not explicitly stated in the original article but can be inferred or deduced. \\
\midrule
\multicolumn{2}{c}{
\cellcolor[rgb]{ .915,  .915,  .915} \textit{\textbf{Layer 3} Decomposition}}  \\
\textit{vocab}  &  Vocabulary: how well the summary uses appropriate vocabulary and expressions, without mis-spelling. \\
\textit{syntax}  &  Syntax: how well the summary uses appropriate sentence structure and word order. \\
\textit{punc}  &  Punctuation: how well the summary uses appropriate punctuation. \\
\textit{len}  &  Length and form: how well the summary is of appropriate length and form to encourage the readers, without being too brief of overly redundant. \\
\textit{smooth}  &  Smoothness: how well the summary is smooth and natural to read, without awkward expressions. \\
\textit{logic}  &  Logic: how well the summary is logical and coherent, without abrupt changes in topic or meaning. A good summary should accurately reflect the logical structure of the original article. \\
\textit{form}  &  Form and genre: how well the summary is of appropriate form and genre to encourage the readers, without being a stack of bullet points. \\
\textit{clarity}  &  Clarity: how well the summary is clear and easy to understand, without ambiguity or confusion. \\
\textit{nat}  &  Naturalness: how well the summary is natural and fluent to read, without awkward transitions or wording. \\


\bottomrule
\end{tabular}}
\caption{A complete case study for criteria decomposition on SummEval.}
\label{tab:decmp-case-2}
\end{table*}
\begin{table*}[ht]
\centering
\footnotesize
\resizebox{0.98\textwidth}{!}{
\begin{tabular}{cp{0.96\textwidth}}
\toprule
Criteria & \textbf{Criteria Decomposition and Definition} \\ 
\midrule
\multicolumn{2}{c}{
\cellcolor[rgb]{ .915,  .915,  .915} \textit{\textbf{Layer 2} Decomposition}} \\
\textit{cov}  &  Coverage: how well the text includes all or most of the important information from the data experssion. \\
\textit{prec}  &  Precision: how accurate and faithful is the text to the data expression. \\
\textit{rel}  &  Relevance: how relevant and salient is the information in the text to the data expression. \\
\textit{gram}  &  Grammaticality: How well does the text follow the rules of grammar and syntax? \\
\textit{read}  &  Readability: How easy is it to read and understand the text? \\
\textit{sty}  &  Style: How well does the text follow the style of the data expression? \\
\midrule
\multicolumn{2}{c}{
\cellcolor[rgb]{ .915,  .915,  .915} \textit{\textbf{Layer 3} Decomposition}}  \\
\textit{datacmp}  &  Data completeness: The proportion of data elements that are mentioned in the text. \\
\textit{datacrr}  &  Data correctness: The accuracy of the information in the text compared to the data. \\
\textit{datared}  &  Data redundancy: The absence of repeated or unnecessary information in the text. \\
\textit{lec}  &  Lexical correctness: The appropriateness and diversity of the words and phrases used in the text. \\
\textit{num}  &  Numerical correctness: The clarity and accuracy of the numerical values and units in the text. \\
\textit{ref}  &  Reference correctness: The accuracy and consistency of the references to entities in the text. \\
\textit{contsel}  &  Content selection: The selection and ordering of the most important and relevant information from the data expression. \\
\textit{contorg}  &  Content organization: The coherence and organization of the information in the text. \\
\textit{contadp}  &  Content adaptation: The adaptation of the information in the text to the target audience. \\
\textit{syn}  &  Syntactic correctness: The correctness of the syntactic structure of the text. \\
\textit{punc}  &  Punctuation correctness: The correctness of the punctuation in the text. \\
\textit{clar}  &  Clarity: The simplicity and directness of the language and expressions in the text. \\
\textit{flu}  &  Fluency: The smoothness and naturalness of the flow and rhythm of the text. \\


\bottomrule
\end{tabular}}
\caption{A complete case study for criteria decomposition on Data-to-Text tasks.}
\label{tab:decmp-case-3}
\end{table*}


\begin{figure*}[!t]
\centering
\begin{tcolorbox}[width=1\textwidth, fontupper=\small, colback=blue!2, boxrule=0.9pt] 
\#\# Instructions

You will be given the conversation history between two individuals, its corresponding fact, and one potential response for the next turn in the conversation.

Please evaluate the \textcolor{purple}{[Parent Criteria]} of the given response to the conversation.

Specifically, to evaluate \textcolor{purple}{[Parent Criteria]}, we would like you to score the given response on the following metric:

\textcolor{blue}{[Child Criteria]} : \textcolor{blue}{[Definition of Child Criteria]}

Please return your score on the above metric in the scale of 1 to 5, with 1 being the lowest. \\

\#\# Example

\textcolor{blue}{[Sample to be evaluated]} \\

\#\# Evaluation

Now, please evaluate the \textcolor{purple}{[Parent Criteria]} of the provided response. (on a scale of 1-5, with 1 being the lowest).
Please carefully read the conversation history, corresponding fact, generated response, and evaluate the sentence using the metric \textcolor{blue}{[Child Criteria]}.
Please first return your score, and then provide your reasoning for the score. \\

Score (1-5):
\end{tcolorbox}
\caption{Hierarchy-Aware Evaluation Prompts for Topical-Chat.}
\label{fig:prompt-chat}
\end{figure*}


\begin{figure*}[!t]
\centering
\begin{tcolorbox}[width=1\textwidth, fontupper=\small, colback=blue!2, boxrule=0.9pt] 
\#\# Instructions

We would like to score the following summary of a news article on its \textcolor{purple}{[Parent Criteria]}.

Specifically, to evaluate \textcolor{purple}{[Parent Criteria]}, we would like you to score the given response on the following metric:

\textcolor{blue}{[Child Criteria]} : \textcolor{blue}{[Definition of Child Criteria]}

Please return your score on the above metric in the scale of 1 to 5, with 1 being the lowest. \\

\#\# Example

\textcolor{blue}{[Sample to be evaluated]} \\

\#\# Evaluation

Now, please evaluate the \textcolor{purple}{[Parent Criteria]} of the provided response. (on a scale of 1-5, with 1 being the lowest).
Please carefully read the conversation history, corresponding fact, generated response, and evaluate the sentence using the metric \textcolor{blue}{[Child Criteria]}.
Please first return your score, and then provide your reasoning for the score. \\

Score (1-5):
\end{tcolorbox}
\caption{Hierarchy-Aware Evaluation Prompts for SummEval.}
\label{fig:prompt-summ}
\end{figure*}


\begin{figure*}[t]
\centering
\begin{tcolorbox}[width=1\textwidth, fontupper=\small, colback=blue!2, boxrule=0.9pt] 
\#\# Instructions

We would like to evaluate the \textcolor{purple}{[Parent Criteria]} of data-to-text, a natural language sentence generated according to a structured data expression.

Specifically, to evaluate \textcolor{purple}{[Parent Criteria]}, we would like you to score the given response on the following metric:

\textcolor{blue}{[Child Criteria]} : \textcolor{blue}{[Definition of Child Criteria]}

Please return your score on the above metric in the scale of 1 to 5, with 1 being the lowest. \\

\#\# Example

\textcolor{blue}{[Sample to be evaluated]} \\

\#\# Evaluation

Now, please evaluate the \textcolor{purple}{[Parent Criteria]} of the provided response. (on a scale of 1-5, with 1 being the lowest).
Please carefully read the conversation history, corresponding fact, generated response, and evaluate the sentence using the metric \textcolor{blue}{[Child Criteria]}.
Please first return your score, and then provide your reasoning for the score. \\

Score (1-5):
\end{tcolorbox}
\caption{Hierarchy-Aware Evaluation Prompts for Data-to-text tasks.}
\label{fig:prompt-d2t}
\end{figure*}

\end{document}